\documentclass{article} %
\usepackage{iclr2027_conference,times}

\usepackage{amsmath,amsfonts,bm}

\def\eqref#1{equation~\ref{#1}}

\def\1{\bm{1}}

\DeclareMathAlphabet{\mathsfit}{\encodingdefault}{\sfdefault}{m}{sl}
\SetMathAlphabet{\mathsfit}{bold}{\encodingdefault}{\sfdefault}{bx}{n}

\usepackage{hyperref}
\usepackage{url}

\usepackage{xspace}
\usepackage{latexsym}
\usepackage{tcolorbox}
\tcbuselibrary{skins, breakable, listings} 
\usepackage{fvextra}
\DefineVerbatimEnvironment{VerbatimWrap}{Verbatim}{
breaklines=true, 
breakindent=0pt, 
breaksymbol={},
fontsize=\small,
}

\usepackage[T1]{fontenc}
\usepackage[utf8]{inputenc}
\usepackage{microtype}
\usepackage{inconsolata}
\usepackage{graphicx}
\usepackage{amsmath}
\usepackage{booktabs}
\usepackage{float}

\usepackage{subcaption}
\usepackage{array}
\usepackage{multirow}

\DeclareCaptionLabelFormat{fullsubtable}{Table~\thetable#2}

\newcommand{\sdtwo}{Seedance 2.5\xspace}
\newcommand{\minimax}{MiniMax H3\xspace}
\newcommand{\sdone}{Seedance 2.0\xspace}
\newcommand{\wan}{Wan 2.2\xspace} 
\newcommand{\hunyuan}{HunyuanVideo 1.5\xspace}
\newcommand{\happyhorse}{HappyHorse 1.0\xspace}
\newcommand{\helios}{Helios-Base\xspace}
\newcommand{\longcat}{LongCat-Video\xspace}
\newcommand{\longlive}{LongLive 1.0\xspace}
\newcommand{\gemini}{Gemini 3.1 Pro\xspace}
\newcommand{\gpt}{GPT-5.4\xspace}

\newcommand{\vbench}{VBench-Long\xspace}
\newcommand{\kivi}{KIVI\xspace}
\newcommand{\kivibench}{KIVI-Bench\xspace}
\newcommand{\fp}{FactP\xspace}
\newcommand{\hs}{HelpS\xspace}

\title{Knowledge-Intensive Video Generation}

\author{Chenxu Wang \\
  Fudan University\\
  \texttt{chenxuwang22@m.fudan.edu.cn} \\\And
  Mingda Chen \\
  Shanghai Jiao Tong University\\
  \texttt{mingdachen@sjtu.edu.cn} \\}

\iclrfinalcopy
\begin{document}

\maketitle
\fancyhead{}
\begin{abstract}

Text-to-video generation has advanced rapidly in visual quality, but existing task settings largely assume that the prompt already specifies what the video should depict. This assumption does not hold in information-seeking scenarios, where users typically specify what they want to learn rather than the detailed visual content of the answer. We introduce \emph{knowledge-intensive video generation} (\kivi \footnote{The code and data are available at \url{https://github.com/wcxhimself/KIVI}}), where models must determine and generate appropriate visual content from concise information-seeking prompts that request explanations, procedures, or demonstrations. Many such requests naturally require long videos that organize multiple factual or procedural details over time. To study this capability, we construct \kivibench, a benchmark of 1,080 prompts targeting long-form video generation, and propose automatic metrics for factuality and helpfulness. Human evaluation shows that our metrics achieve substantially stronger agreement with human judgments than existing visual-quality-oriented alternatives. Experiments on nine state-of-the-art video generation models reveal substantial variation in \kivi performance, while even the strongest models exhibit frequent failures in entity-specific visual properties, procedural operations, and clear information presentation. These results highlight the need to move beyond visually plausible rendering toward video generation systems that can reliably communicate correct and useful information.

\end{abstract}

\section{Introduction}

Text-to-video generation has advanced rapidly in visual quality, driven by recent progress in generative modeling \citep{ho2022imagen,singer2023makeavideo}. As these models become increasingly capable and accessible, their potential is expanding beyond entertainment-oriented generation to applications such as healthcare \citep{info:doi/10.2196/88005}, education \citep{10.1007/978-3-031-36336-8_81,educsci15010102}, and live Q\&A \citep{klein2024delivering}. These applications introduce a different type of user request. Instead of describing the video they want to see, users may simply ask how to perform a procedure, understand a concept, or use a particular object.

Current text-to-video task settings are built around a different type of input. The prompt typically specifies the desired scenes, objects, and actions, and the model is mainly responsible for realizing them visually. Accordingly, existing benchmarks emphasize visual quality, temporal consistency, and motion smoothness \citep{wu2024towards,DBLP:conf/cvpr/HuangHYZS0Z0JCW24}. This setup is less suitable in the information-seeking scenarios where users do not know in advance what the video should contain.

\begin{figure}[h] 
\centering
\includegraphics[width=\textwidth]{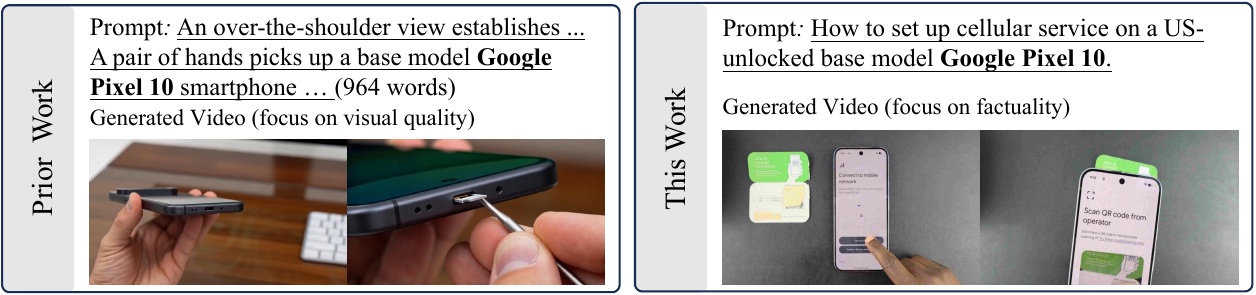} 
\caption{
Visual-quality-driven (left) vs. knowledge-intensive (right) video generation. Prior work typically assumes detailed prompts that specify what should be shown, whereas our setting starts from short information-seeking prompts that specify what the user wants to learn. In this example, the visually plausible left video incorrectly depicts a physical SIM-card setup, while the right video correctly shows software-based cellular setup.}
\label{fig:intro}
\end{figure}

Figure~\ref{fig:intro} illustrates this issue. The user asks how to set up cellular service on a Google Pixel 10, without specifying the details that should appear in the video. The left video is visually convincing but incorrectly depicts a physical SIM-card setup, while the right video correctly shows software-based setup. A model can therefore satisfy conventional notions of visual quality while still giving the user the wrong answer. Thus, the task goal is no longer only to generate visually plausible content, but to communicate information that is both factually correct and useful to the user.

We refer to this setting as knowledge-intensive video generation (\kivi), where models must determine and generate appropriate visual content from concise information-seeking prompts. \kivi is closely related to information-seeking long-form text generation \citep{NEURIPS2020_1457c0d6,lee2022factuality}, but differs in the knowledge required. Text generation can often rely on knowledge expressed in words, while generating factually correct videos depends more heavily on visual, spatial, and procedural knowledge that is rarely verbalized. For example, models need to know what an object looks like, where its components are located, and how an action unfolds over time.

Many such requests also naturally call for long videos, as a useful answer may need to present multiple factual details, procedural steps, and intermediate states in a coherent sequence. This makes knowledge-intensive generation particularly challenging: the system needs not only to determine the relevant information, but also to organize and communicate it effectively over time. To study this capability, we construct \kivibench, a benchmark of 1,080 concise information-seeking prompts targeting at long-form video generation and covering diverse everyday and professional scenarios. 

We further propose two complementary automatic metrics. The first measures factuality by estimating the fraction of verifiable claims conveyed in the generated video that are factually correct, following the intuition of claim-level factual precision in long-form text generation \citep{manakul-etal-2023-selfcheckgpt,min-etal-2023-factscore}. The second measures helpfulness, capturing whether the video provides information useful for satisfying the user's request. Together, these metrics evaluate whether a generated video is both factually reliable and practically useful.

We validate our metrics through human evaluation and find that they achieve stronger agreement with human judgments than existing evaluation alternatives for both factuality and helpfulness. We then benchmark nine state-of-the-art video generation models, including both closed-source and open-source systems. Performance varies substantially across models, and even the strongest systems exhibit persistent failures in entity-specific visual properties, procedural operations, and clear information presentation. These results highlight the need for video generation systems that can more reliably communicate factual and instructionally useful information.

To summarize, our contributions are as follows.
\begin{itemize}
    \item We formulate \kivi as a new task setting for evaluating text-to-video generation beyond visual quality, where models generate videos from short information-seeking prompts rather than fully specified scene descriptions.
    \item We construct \kivibench, a benchmark of 1,080 knowledge-intensive prompts covering diverse instructional and information-seeking scenarios.
    \item We introduce automatic factuality and helpfulness metrics for generated videos, achieving relative gains of 25.3\% and 30.7\% in agreement with human judgments over the best visual-quality baselines, respectively.
    \item We benchmark nine state-of-the-art video generation models and show that \kivibench remains challenging for current methods, with detailed analyses of common model failures.
\end{itemize}

\section{Related Work}

\paragraph{Text-to-Video Generation.}
Recent work on text-to-video generation has made rapid progress by extending diffusion models, transformer-based generative models, and large-scale multi-modal pretraining to the video domain \citep{ho2022imagen,hong2023cogvideo,singer2023makeavideo,villegas2023phenaki,DBLP:conf/cvpr/BlattmannRLD0FK23,pmlr-v235-kondratyuk24a}. Alongside model development, several benchmarks have been proposed to evaluate generated videos along dimensions such as visual quality, temporal consistency, and motion smoothness \citep{liu2023fetv,DBLP:conf/cvpr/LiuC0WZCLZCS24,DBLP:conf/cvpr/HuangHYZS0Z0JCW24,DBLP:conf/cvpr/SunHL0XLL25}. More recent studies further examine whether generated videos obey physical commonsense \citep{bansal2025videophy,meng2025towards,bansal2026videophy}. More closely related to our work, \citet{wang2025respond} evaluate video generation models using real user queries involving knowledge explanation, art creation, and human-machine interaction, while \citet{Chen2026T2VWorldBenchAB} and \citet{wang2026videoverse} introduce benchmarks for evaluating world knowledge in traditional text-to-video generation, with an emphasis on cultural and physical plausibility. In contrast, our work formulates knowledge-intensive video generation as a realistic evaluation setting and studies whether generated videos are both factually accurate and helpful across a broad range of knowledge-seeking prompts.

\paragraph{Multi-Modal Knowledge-Intensive Tasks.}
A large body of work has studied multi-modal tasks that require external knowledge beyond the visual input. Knowledge-based visual question answering benchmarks such as OK-VQA \citep{DBLP:conf/cvpr/MarinoRFM19}, KVQA \citep{10.1609/aaai.v33i01.33018876}, and A-OKVQA \citep{10.1007/978-3-031-20074-8_9} require models to answer questions using commonsense, encyclopedic, or structured world knowledge in addition to image understanding. More recent benchmarks further emphasize information-seeking and fine-grained knowledge, such as InfoSeek \citep{chen-etal-2023-pre-trained} and Encyclopedic-VQA \citep{10376881}, where questions often require recognizing visual entities and retrieving relevant external knowledge \citep{10376881}. Other work extends knowledge-intensive reasoning to heterogeneous multi-modal contexts, including text, tables, and images \citep{talmor2021multimodalqa}. More recent work has extended the setups to videos \citep{DBLP:conf/aaai/GarciaOCN20,he2025mmworld,DBLP:conf/cvpr/00010XHGLHCLXWS25,DBLP:conf/aaai/CaoHWGTZWDYZLRL26}. However, they primarily evaluate understanding and question answering over given multi-modal evidence. Our work studies a complementary generation problem: instead of answering questions about existing visual content, models must generate videos that accurately convey the knowledge requested by a textual prompt.

\begin{figure*}[t]  
\centering
\includegraphics[width=\textwidth]{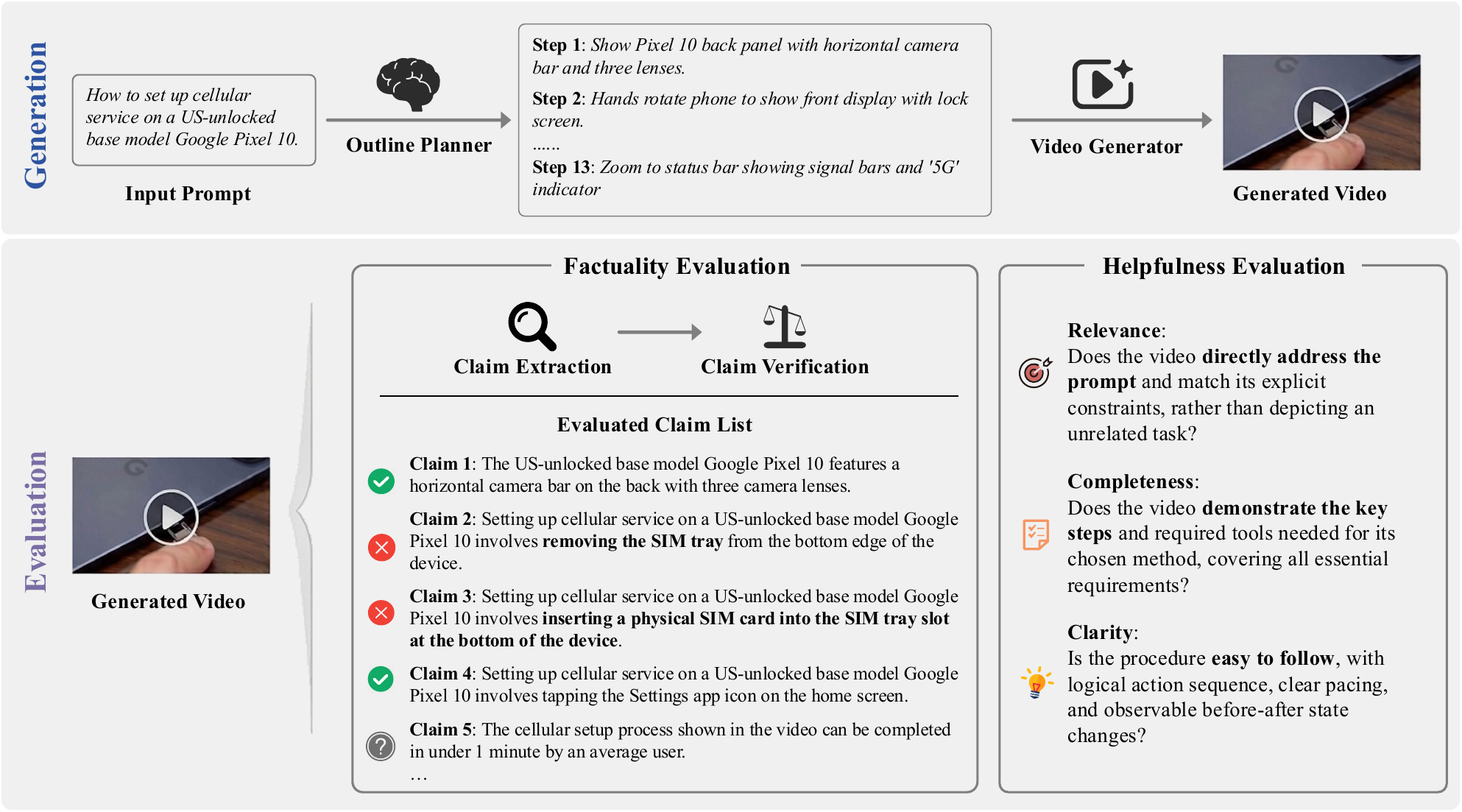} 
\caption{A diagram illustrating the generation and evaluation pipeline used in this work. During generation, the input prompt is first passed to an LLM to produce a multi-step outline of the desired video, which is then used by a video generator to produce the final video. During evaluation, the generated video is assessed with LLM-based metrics along two aspects: factuality and helpfulness. Factuality measures the fraction of correct atomic claims conveyed in the video, while helpfulness measures how well the video satisfies the user request along three dimensions. In the shown example, the incorrect claims state that the Google Pixel 10 uses physical SIM cards.}
\label{fig:overall_pipeline}
\end{figure*}

\paragraph{Multi-Modal Fact-Checking.}
Fact-checking has traditionally focused on verifying textual claims against textual evidence \citep{thorne-etal-2018-fever}, but recent work has extended the setting to multi-modal evidence and misinformation. Multi-modal fact-checking datasets and systems verify claims using both text and images, often requiring models to retrieve evidence, detect cross-modal inconsistencies, and predict whether a claim is supported or refuted \citep{chakraborty-etal-2023-factify3m}. Another line of work studies out-of-context or mismatched image-text misinformation, where both the image and caption may be individually real but misleading when paired together \citep{luo-etal-2021-newsclippings,DBLP:conf/cvpr/AbdelnabiHF22,tonglet-etal-2025-cove}. These works highlight the importance of verifying cross-modal factual consistency. However, they typically assume that the claim, evidence, or image-text pair is already given. In contrast, knowledge-intensive video generation requires evaluating factuality in generated videos, where the relevant claims must first be inferred from the video content and then checked against external knowledge. Our work therefore connects text-to-video evaluation with multi-modal fact-checking, but targets the distinct problem of measuring whether generated videos faithfully and helpfully communicate correct information.

\section{Knowledge-Intensive Video Generation}

Knowledge-intensive video generation aims to generate factually accurate and useful videos in response to concise information-seeking prompts. The overall generation and evaluation pipeline is shown in Figure~\ref{fig:overall_pipeline}. In the following subsections, we describe how the prompt set and evaluation metrics are constructed.

\subsection{KIVI-Bench Construction} \label{sec:kivibench} 
The prompt set, which we refer to as KIVI-Bench, is constructed through the following pipeline.
\paragraph{Prompt Topic Selection.}
To ensure diverse topic coverage,
we use 18 topic categories from WikiHow Video\footnote{\url{https://www.wikihow.com/Videos}}, covering everyday life and professional domains. See Appendix~\ref{app:prompt_topics} for details on the topics.

\paragraph{Prompt Set Generation.}
For each category, we manually construct several seed prompts that satisfy five criteria: (1) video demonstration is more suitable than text explanation; (2) the prompt has a valid premise and an answer verifiable against readily available documentation; (3) the prompt specifies distinctive, uniquely identifiable entities rather than generic objects; (4) answering it requires non-obvious factual or procedural knowledge not provided in the prompt; and (5) it is concise and natural, resembling a real-world user query. Using these seeds as in-context examples and the template in Figure~\ref{app:prompt_generation_template}, we ask \gpt \citep{openai2026gpt54} to generate 80 prompts per category, yielding 1,440 candidates.

\paragraph{Quality Control.}
We conduct a two-stage review of the 1,440 candidates. First, \gpt flags potentially duplicate prompts across the full candidate set based on their target entities and requested tasks (See Figure~\ref{app:dedup_prompt}). A human reviewer then merges, retains, or removes the flagged prompts, reducing the set to 1,152 candidates. Second, a human reviewer removes prompts with ambiguous entity references, insufficient reliable documentation, or no well-established, verifiable answer. This stage removes 72 candidates, resulting in the final set of 1,080 prompts.

\subsection{\kivibench Evaluation Metrics} \label{sec:metric} 
To support automatic evaluation, we propose two LLM-based metrics that assess complementary aspects of generated videos.

\paragraph{Factual Precision (\fp).} Inspired by factuality evaluation in long-form text generation \citep{manakul-etal-2023-selfcheckgpt,min-etal-2023-factscore}, we design an LLM-based metric that first reviews the generated video and extracts \emph{video claims}. We define a video claim as an atomic, externally verifiable factual statement about what the video visually depicts. Claims are extracted by an LLM using the prompt in Figure~\ref{app:extraction_prompt}.\footnote{Although multi-modal claims that combine visual and textual information are more natural than text-only claims, our analysis shows that they are not straightforward to implement (See Section ~\ref{sec:multimodal} for more details). We therefore use text-only claims throughout our experiments and leave a more thorough exploration of multi-modal claims to future work.} Each extracted claim is then verified against world knowledge\footnote{We note that while retrieval from external sources is a common approach for accessing world knowledge, in this work we rely on the LLM's parametric knowledge to verify claim factuality. This choice is motivated partly by prior work in the text domain showing that LLM-based verification can be feasible \citep{manakul-etal-2023-selfcheckgpt}, and partly by the limitations of existing retrieval sources: they are typically text-based and are not well suited to our setting, where the relevant information is often multimodal and absent from text-only knowledge corpora.} and classified as ``Correct'', ``Incorrect'' or ``Uncertain'' using the verification prompt in Figure~\ref{app:verification_prompt}. For each data item, \fp is defined as:

\[
\text{\fp} = \frac{\text{number of correct claims}}{\text{total number of claims}} \times 100\%
\]

We then average \fp across all data items to obtain the dataset-level metric score.

\paragraph{Helpfulness Score (\hs).} While \fp measures the precision of factual claims, it does not capture whether the video adequately satisfies the user's request. We therefore introduce a recall-oriented helpfulness score. Given a video, an LLM reviews the video and rates it along three dimensions (See the prompt in Figure~\ref{app:helpfulness_prompt}), each on a scale from 0 to 10: \emph{Relevance}, measuring whether the video addresses the user request; \emph{Completeness}, measuring whether key steps or information are covered; and \emph{Clarity}, measuring whether the content is easy to follow. The final score is computed as:\footnote{While more effective weighting schemes may exist, we find that simple averaging gives reasonable performance and leave further exploration to future work.}
\[
\text{\hs} = \frac{\text{Relevance} + \text{Completeness} + \text{Clarity}}{3} \times 100\%
\]

Similar to \fp, we report the dataset-level \hs by averaging over all data items.

\section{Experiments} \label{sec:experiment}

Due to computational and budget constraints, all experiments use a uniformly sampled subset of 54 prompts, with 3 prompts from each category. All LLM calls for video planning, script generation, and automatic evaluation use \gemini \citep{gemini31pro2026} with temperature 0. We choose \gemini because of its strong video-understanding capability. More details on computational resources are reported in Appendix~\ref{app:comp_resource}.

\subsection{Evaluation Setup} \label{sec:setup}
We evaluate nine state-of-the-art video generation models on \kivibench, including
\begin{itemize}
    \item Three closed-source API models: \sdtwo\footnote{\url{https://seed.bytedance.com/zh/seedance2_5}}, \sdone \citep{seedance2026seedance20advancingvideo} and \happyhorse\footnote{\url{https://www.aliyun.com/benefit/scene/happyhorse}}
    \item Six open-source models: \minimax\footnote{\url{https://www.minimax.io/}}, \wan~(A14B; \citealp{DBLP:journals/corr/abs-2503-20314}),  \hunyuan \citep{wu2025hunyuanvideo15technicalreport},  \helios \citep{yuan2026heliosrealrealtimelong}, \longlive \citep{yang2026longlive} and \longcat \citep{meituanlongcatteam2025longcatvideotechnicalreport}
\end{itemize}
Each model is prompted to generate videos of approximately 60 seconds. For all evaluations on \kivibench, we use the official implementation of each model\footnote{Due to budget constraints, the Seedance models (\sdtwo{} and
  \sdone{}) are generated at 480p resolution, while all other models use 720p.}.

The long video generation models (\helios, \longlive, and \longcat) support two modes: interactive and single-prompt generation. In interactive mode, the model receives a sequence of time-stamped sub-prompts, each describing one stage of the procedure; in single-prompt mode, it generates the full video from a single prompt. In preliminary experiments, we find that interactive mode generally produces richer and more temporally structured videos. We therefore use interactive mode for all long video models except \helios, for which we use single-prompt mode due to technical issues in its official implementation.

For short video generation models (\sdtwo, \sdone, \happyhorse, \minimax, \wan, and \hunyuan), which produce single video clips with a maximum duration of 30 seconds, we adopt a similar interactive pipeline. Specifically, we first use an LLM to convert the input prompt into a multi-step visual outline. The first clip is generated from the textual outline of the initial step. Each subsequent clip is generated by conditioning on both the last frame of the previous clip and the textual outline of the current step. We then stitch all clips together to obtain the final long video. Appendix~\ref{app:interact_vs_single_ablation} provides more detailed comparisons between interactive and single-prompt generation for these models.

In addition to our factuality and helpfulness metrics, we evaluate generated videos using six quality dimensions from \vbench \citep{DBLP:journals/pami/HuangZXHYDMCSJWCCWLQL26}: motion smoothness, imaging quality, dynamic degree, aesthetic quality, subject consistency, and background consistency. These metrics serve as visual quality baseline metrics for comparison with our proposed evaluation metrics.

\begin{table}[t]
    \centering
    \caption{Overall results on the subset of \kivibench. \#Cor. = number of correct claims, \#Inc. = number of incorrect claims, \#Unc. = number of uncertain claims. \fp = Factual Precision. Rel = Relevance. Cmp = Completeness. Clr = Clarity. \hs = Helpfulness Score. The best performance in each column is boldfaced.}
    \label{tab:main_result}
    \resizebox{\textwidth}{!}{%
    \footnotesize           
    \setlength{\tabcolsep}{3pt}
    \begin{tabular}{lccccc|cccc}
        \toprule
        & \multicolumn{5}{c|}{\textbf{Factuality}} & \multicolumn{4}{c}{\textbf{Helpfulness}} \\
        \textbf{Model} & \textbf{\#Cor.} ($\uparrow$) & \textbf{\#Inc.} ($\downarrow$) & \textbf{\#Unc.} ($\downarrow$) & \textbf{Total} ($\uparrow$) & \textbf{\fp} (\%, $\uparrow$) & \textbf{Rel} ($\uparrow$) & \textbf{Cmp} ($\uparrow$) & \textbf{Clr} ($\uparrow$) & \textbf{\hs}(\%, $\uparrow$) \\
        \midrule
        \multicolumn{10}{l}{\emph{Closed-source short video generation models}}\\
        \sdtwo   & 396 & 78  & 12  & 486 & 82.3 & 77.2 & 70.9 & 61.7 & 69.9 \\
        \sdone     & 367 & 77  & 6  & 450 & 81.6 & 75.7 & 69.8 & 54.1 & 66.6 \\
        \happyhorse   & 362 & \textbf{66}  & 9  & 437 & 83.2 & 70.2 & 66.5 & 48.1 & 61.6 \\
        \cmidrule{1-10}
        \multicolumn{10}{l}{\emph{Open-source short video generation models}}\\
        \minimax   & \textbf{442} & \textbf{66}  & 6  & \textbf{514} & \textbf{86.8} & \textbf{81.5} & \textbf{76.3} & \textbf{70.9} & \textbf{76.2} \\
        \wan         & 306 & 112 & 8  & 426 & 73.1 & 57.4 & 53.0 & 34.8 & 48.4 \\
        \hunyuan & 259 & 139 & \textbf{5}  & 403 & 63.2 & 41.3 & 38.0 & 19.3 & 32.9 \\
        \cmidrule{1-10}
        \multicolumn{10}{l}{\emph{Open-source long video generation models}}\\
        \helios           & 221 & 123 & 6  & 350 & 64.2 & 43.3 & 24.1 & 13.5 & 27.0 \\
        \longcat    & 184 & 152 & \textbf{5}  & 341 & 50.8 & 23.3 & 15.4 & 7.2  & 15.3 \\
        \longlive         & 176 & 201 & 7  & 384 & 46.5 & 28.3 & 15.4 & 3.9  & 15.9 \\
        \bottomrule
    \end{tabular}}
\end{table}

\begin{table}[t]
    \centering
    \caption{Human evaluation results. Annotators choose their preferred video from each presented pair based on factuality and helpfulness. Agreement approximately measures the fraction of decisions where the metric-preferred video matches the human preference.}
    \label{tab:human_eval}
    \small
    \begin{tabular}{lcc}
        \toprule
        \textbf{Metric} & \textbf{Fact. (\%)} & \textbf{Help. (\%)} \\
        \midrule
        \multicolumn{3}{l}{\emph{Prior evaluation metrics}} \\
        Motion Smoothness    & 39.8 & 44.4 \\
        Imaging Quality      & 38.9 & 38.0 \\
        Dynamic Degree       & 56.5 & 52.8 \\
        Aesthetic Quality    & 44.4 & 52.8 \\
        Subject Consistency  & 44.4 & 45.4 \\
        Background Consistency & 51.9 & 52.8 \\
        Overall Visual Quality & 48.1 & 47.2 \\
        \midrule
        \multicolumn{3}{l}{\emph{Our proposed metrics}} \\
        Factual Precision          & \textbf{70.8} & -- \\
        Helpfulness Score          & -- & \textbf{69.0} \\
        \bottomrule
    \end{tabular}
\end{table}

\subsection{Benchmarking on \kivibench}

Table~\ref{tab:main_result} presents the overall results. \minimax{} achieves the highest \fp{} and \hs{}, outperforming all closed-source models on both metrics. Among closed-source models, \happyhorse{} attains the highest \fp{}, while \sdtwo{} achieves the highest \hs{}. Among open-source models, \wan{} ranks second on both metrics. The long-video models (\helios, \longcat, and \longlive) perform substantially worse, with \fp{} ranging from 46.5\% to 64.2\% and \hs{} below those of all evaluated short-video models.

Performance also varies considerably across models, with a 40-point gap in \fp{} and a 61-point gap in \hs{} between the best and worst systems. Among the helpfulness dimensions, Clarity is consistently the most challenging, receiving the lowest score for every model and ranging from 3.9 to 70.9. This suggests that producing clear and followable demonstrations remains difficult, particularly when factual errors disrupt the progression of a procedure. Even the strongest model achieves only slightly above 86\% \fp{}, and this score may still be optimistic: as discussed in Section~\ref{sec:eval_coverage}, subtle visual errors can be missed during claim extraction or judged as correct during verification.

Notably, \vbench produces a substantially different model ranking. The long-video models, despite ranking lowest on \fp{} and \hs{}, score highest on several visual-quality dimensions. This divergence suggests that these dimensions primarily capture visual stability and quality rather than a model's ability to communicate factual and useful information. Full results are reported in Appendix~\ref{app:vbench_long_full_results}.

\subsection{Human Evaluation} \label{sec:human}
To validate the effectiveness of our automatic metrics, we conduct a human evaluation study. Annotators were presented with two videos generated for the same prompt and asked to choose which one they preferred in terms of factuality and helpfulness, with the two preferences collected separately. Annotators were encouraged to use online search tools, and we provided links to reference videos, to verify relevant information. Detailed annotation instructions are provided in Appendix~\ref{app:human_eval}.

In total, we constructed 108 pairwise comparisons, each annotated by exactly one of our six annotators for both factuality and helpfulness. To compute human--metric agreement, a metric receives 1 point if it prefers the same model as the human annotator, 0.5 points if the two models are tied according to the metric, and 0 otherwise. The final agreement score is obtained by averaging over all comparisons.

Table~\ref{tab:human_eval} reports the results. Under this protocol, our \fp achieves 70.8\% agreement with human factuality judgments, outperforming the best \vbench dimension (56.5\%) by a relative gain of 25.3\%. Our Helpfulness Score achieves 69.0\% agreement, surpassing the best \vbench dimension (52.8\%) by a relative gain of 30.7\%. In contrast, the weakest \vbench dimensions, Imaging Quality and Motion Smoothness, agree with human judgments only 38.9\% and 39.8\% of the time. \vbench Overall also trails our metrics by more than 20 points on both dimensions. These results show that our LLM-based metrics better capture human-perceived factual accuracy and utility than traditional visual-quality-oriented metrics.

\section{Analysis} \label{sec:analysis}

In this section, we analyze the difficulty of our task setup and the characteristics of our evaluation pipeline. Category-level results show that models perform better on topics requiring fewer fine-grained details but struggle more on categories demanding precise factual knowledge (Appendix~\ref{app:cat_perf_analysis}). Our error analysis further reveals persistent weaknesses in entity-specific visual knowledge, procedural knowledge, and component localization (Appendix~\ref{sec:error}).

\subsection{Multi-Modal vs. Text-Only Claim Verification} \label{sec:multimodal}
We compare three claim verification strategies using outputs from four models (\sdone, \happyhorse, \wan, and \hunyuan). \textit{Text-only} verifies each claim using only its textual form. \textit{Text+Video} pairs each claim with a short video clip. \textit{Text+Image} pairs each claim with a key frame, defined as the middle frame of the corresponding video clip. For the multi-modal modes, we keep the textual claims unchanged and provide the visual content only as additional evidence, whose temporal location is identified by \gemini. This design allows the text to preserve broader contextual information that may not be fully captured by a single image or video clip.

\begin{table}[t]
    \centering
    \small
    \setlength{\tabcolsep}{4pt}
    \renewcommand{\arraystretch}{1.2}
    \caption{Human agreement on factuality preferences across three claim verification modes. \textit{Text-only} verifies the textual claim alone; \textit{Text+Image} pairs the claim with a single keyframe; \textit{Text+Video} pairs the claim with a short clip of roughly 2--10 seconds covering the claim's timestamp (three sampled frames shown here). For simplicity, the claim text shared across all three modes is shown only once.}
    \label{tab:claim_modes}
    \begin{tabular}{p{4.0cm} p{3.8cm} c c}
        \toprule
        \multicolumn{2}{c}{\textbf{Example Claim}} & \textbf{Mode} & \textbf{Agree. (\%)} \\
        \cmidrule(lr){1-2}
        \textbf{Text} & \centering\textbf{Visual Evidence} & & \\
        \midrule
        
        \multirow{3}{4.2cm}{\raggedright\footnotesize Setting up cellular service on a US-unlocked base model Google Pixel 10 can involve using a metal ejector tool to open a SIM card tray located on the bottom edge next to the charging port.}
        
        & \centering -- & Text-only & 70.4 \\
        \cmidrule(lr){2-4} 
        
        & \centering \includegraphics[width=1.1cm]{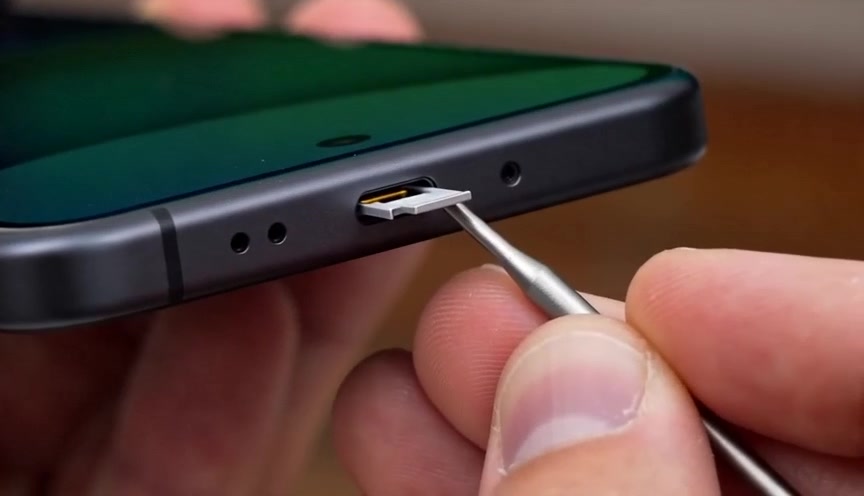} & Text+Image & 58.3 \\
        \cmidrule(lr){2-4}
        
        & \centering 
        \includegraphics[width=1.1cm]{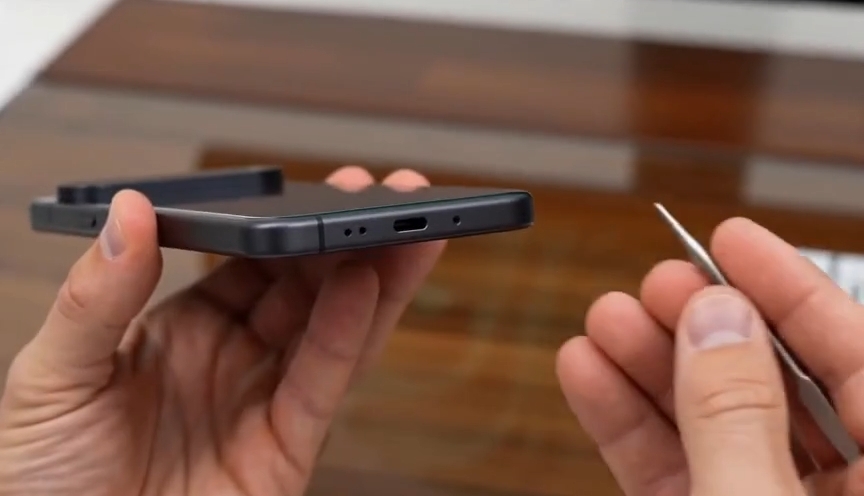}\hspace{1pt}
        \includegraphics[width=1.1cm]{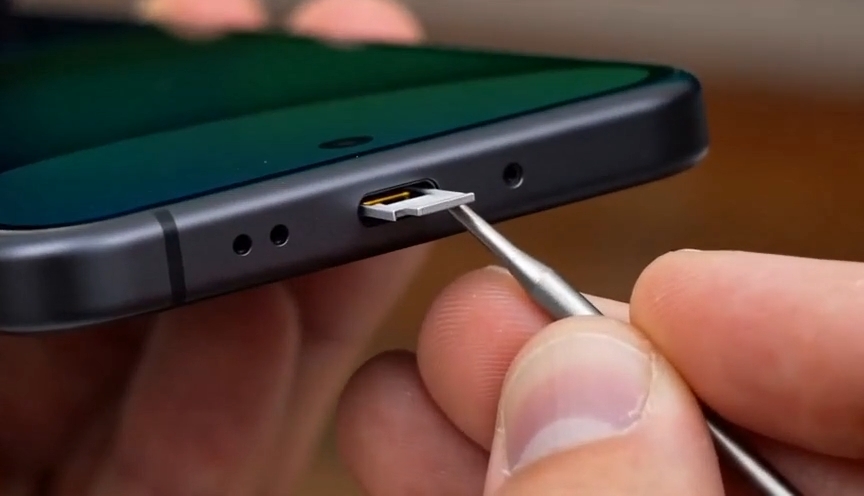}\hspace{1pt}
        \includegraphics[width=1.1cm]{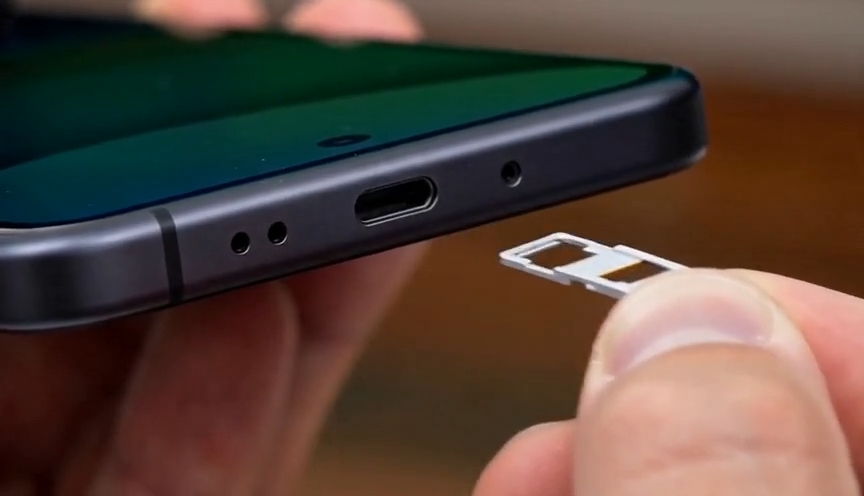} 
        & Text+Video & 58.3 \\
        \bottomrule
    \end{tabular}
\end{table}

We evaluate each verification mode by comparing its factuality preferences with human judgments. As shown in Table~\ref{tab:claim_modes}, Text-only clearly outperforms the two multi-modal variants. We hypothesize that this gap partly arises from a tradeoff introduced by visual evidence. Images and video clips provide broader context around a claim, but they also introduce additional objects, actions, and details that may not align one-to-one with the claim being verified. This added context can therefore weaken the atomicity of the claim and make it harder to isolate the relevant evidence. In comparison, textual claims are easier to represent as atomic units, enabling cleaner and more stable verification. Multi-modal verification also yields substantially lower \fp scores, likely reflecting the same ambiguity in judging multi-modal claims (See Appendix~\ref{app:mm_claim_ablation}).

Text-only claims also have limitations. They may miss visual errors that require grounding actions, objects, or product appearances in the video. For example, linking a depicted action or object to the correct real-world entity can be difficult from text alone. In this work, following prior long-form factuality evaluation, we use proper nouns from the prompt as the primary subjects of textual claims. We leave more explicit visual grounding of claims to future work.

\subsection{Qualitative Analysis of Automatic Claim Extraction}
\label{sec:eval_coverage}

Automatic claim extraction can also misrepresent or underspecify visual details that are important for factual verification. Table~\ref{tab:extraction_error} shows two representative cases. In the Chase check-deposit video generated by \minimax{}, the review screen displays two front-side check images, but the extracted claim incorrectly describes them as the front and back of the check. In the Omron BP5450 video generated by \sdtwo{}, the depicted device differs from the specified model in its display and button layout, but the extracted claim describes only generic components and lacks the detail needed for reliable verification. Neither issue is captured elsewhere in the corresponding claim set. As a result, all 11 claims for the Chase video and all 10 claims for the Omron video are judged correct, yielding an \fp{} of 100\% despite the underlying visual errors.

These examples reveal a gap between the correctness of extracted textual claims and their fidelity to the underlying video. When claim extraction misses or abstracts away important visual details, \fp{} can overestimate factuality even if downstream verification is accurate. A similar issue can affect helpfulness when the evaluator recognizes the presence of a step but fails to detect that it is performed incorrectly. Thus, although both metrics show stronger agreement with human preferences than the evaluated visual-quality baselines (Section~\ref{sec:human}), limitations in visual understanding can still produce overly favorable assessments of factual accuracy and instructional usefulness. Improving claim extraction and evaluation to better capture fine-grained visual details remains an important direction for future work.

\begin{table}[t]
    \centering
    \small
    \setlength{\tabcolsep}{4pt}
    \renewcommand{\arraystretch}{1.05}
    \caption{
Examples of extracted claims that miss or misinterpret important visual details, with the relevant text highlighted. Top (\minimax): two front-side check images are incorrectly described as the front and back of the check. Bottom (\sdtwo): the extracted claim describes the device too generically to support reliable verification of its display and button layout.
    }
    \label{tab:extraction_error}

    \begin{tabular}{
        @{}
        >{\raggedright\arraybackslash}p{0.22\textwidth}
        >{\raggedright\arraybackslash}p{1.8cm}
        >{\raggedright\arraybackslash}p{1.8cm}
        >{\raggedright\arraybackslash}p{
            \dimexpr0.78\textwidth-3.6cm-24pt\relax
        }
        @{}
    }
        \toprule
        \textbf{Prompt}
        & \textbf{Expected\newline Video}
        & \textbf{Generated\newline Video}
        & \textbf{Example Extracted Claim} \\
        \midrule

        \textit{Show how to deposit a check using the
        Chase Mobile app.}
        &
        \begin{minipage}[t]{\linewidth}
            \vspace{0pt}
            \centering
            \includegraphics[width=1.8cm]
                {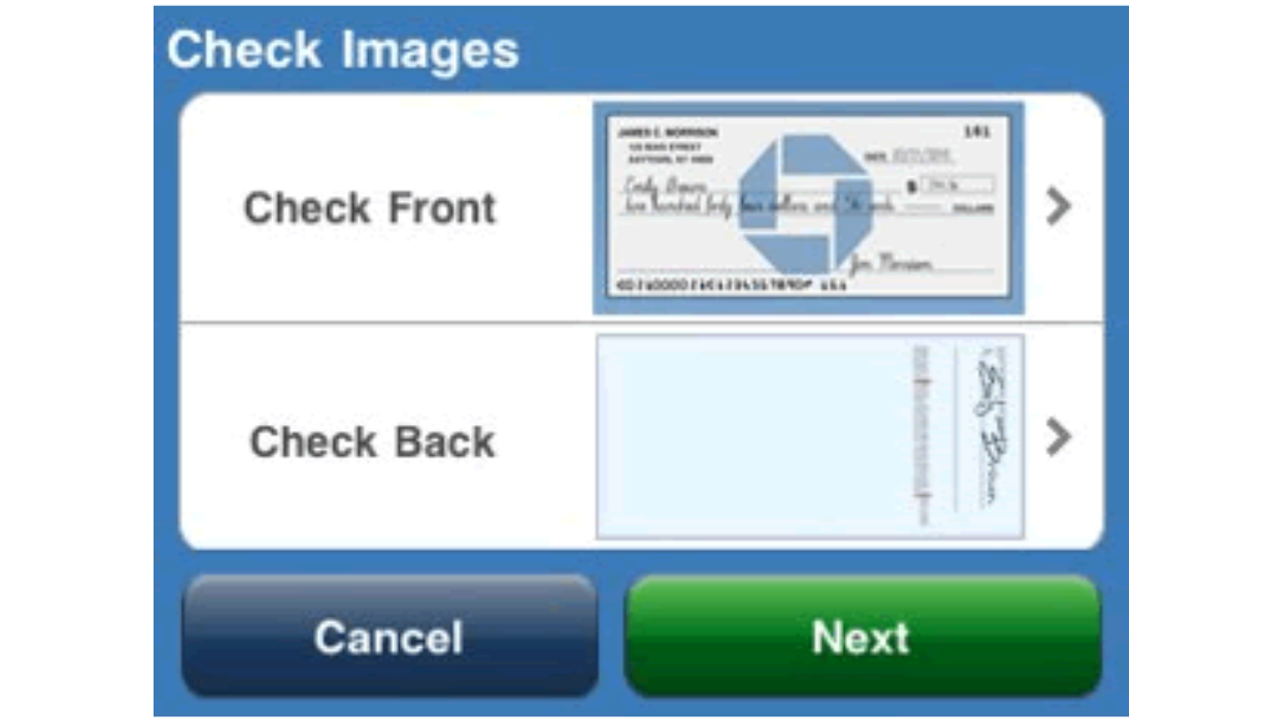}
        \end{minipage}
        &
        \begin{minipage}[t]{\linewidth}
            \vspace{0pt}
            \centering
            \includegraphics[width=1.8cm]
                {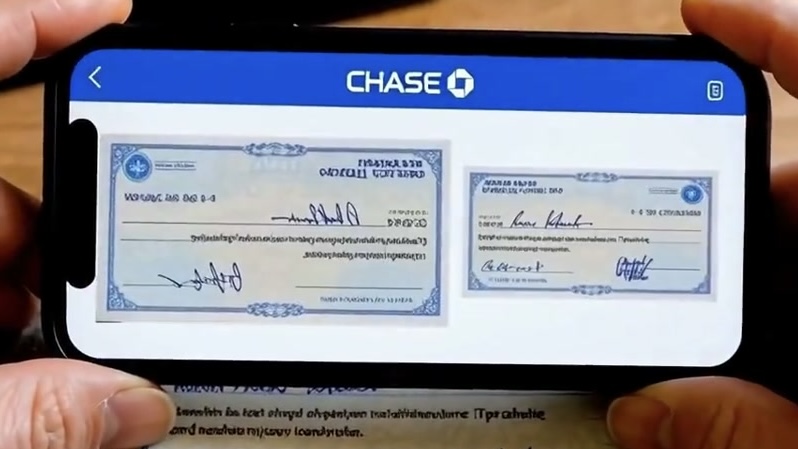}
        \end{minipage}
        &
        \textbf{Claim:}
        ``The Chase Mobile app displays a review screen showing
        the captured images of \textbf{both the front and back
        of the check} before final submission''
        \\
        \midrule

        \textit{How to measure blood pressure using an
        Omron Platinum BP5450.}
        &
        \begin{minipage}[t]{\linewidth}
            \vspace{0pt}
            \centering
            \includegraphics[width=1.8cm]
                {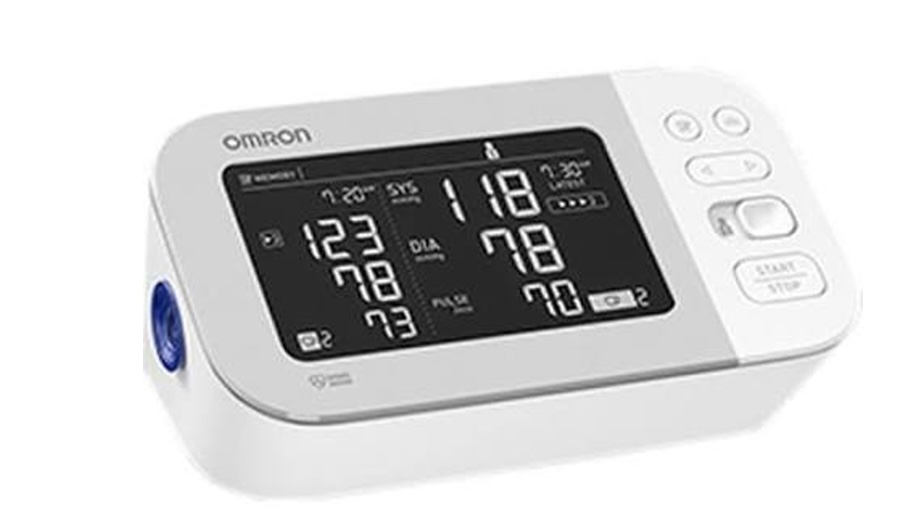}
        \end{minipage}
        &
        \begin{minipage}[t]{\linewidth}
            \vspace{0pt}
            \centering
            \includegraphics[width=1.8cm]
                {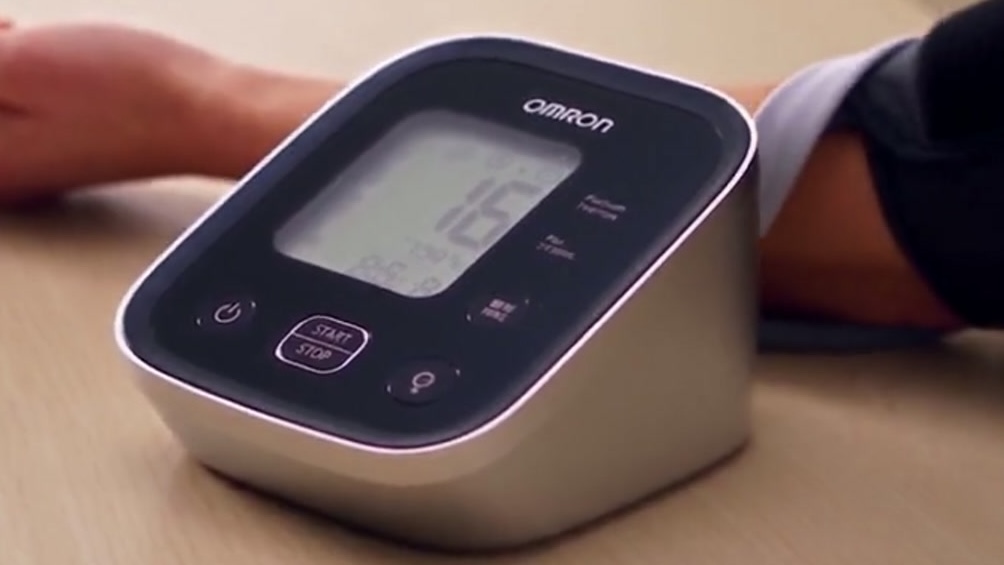}
        \end{minipage}
        &
        \textbf{Claim:}
        ``The \textbf{Omron Platinum BP5450} consists of a
        \textbf{main monitor unit with a digital display}
        and a separate arm cuff attached to an air tube''
        \\
        \bottomrule
    \end{tabular}
\end{table}

\subsection{Impact of Outline Factuality} \label{sec:outline}

\begin{table*}[t]
    \centering
    \caption{Examples where the outline is correct but the generated video is factually incorrect. Inaccuracies may arise from the video generation model rather than the outline.}
    \label{tab:outline_factuality}
    
    \footnotesize
    \setlength{\tabcolsep}{5pt}
    \renewcommand{\arraystretch}{1.15}
    
    \begin{tabular}{p{0.30\textwidth} p{0.42\textwidth} p{0.20\textwidth}}
        \toprule
        \textbf{Prompt} & \textbf{Outline Step} & \textbf{Generated Video} \\
        \midrule
        
        {\raggedright\small
        \textit{Show how to replace the cabin air filter on a Toyota RAV4 (XA50) to improve HVAC airflow.}}
        &
        {\raggedright\small
        Hand opens the \textbf{passenger side door} of a Toyota RAV4 (XA50), revealing the dashboard and closed glovebox.
        \newline{\tiny (Step 1)}}
        &
        \begin{minipage}[t]{\linewidth}
            \vspace{0pt}
            \centering
            \includegraphics[width=1.8cm]{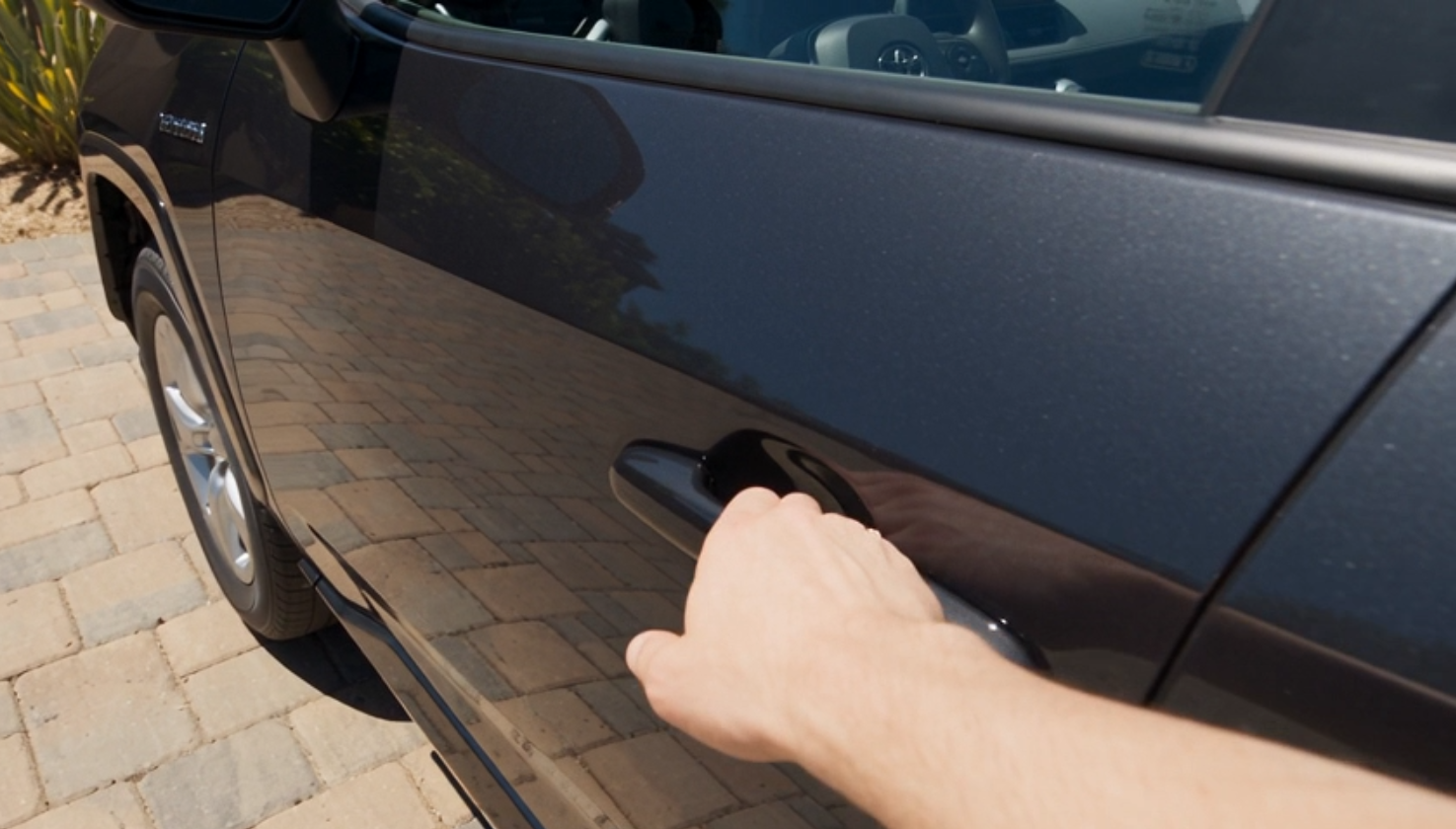}
            \newline
            \scriptsize\sdtwo
        \end{minipage}
        \\
        
        \cmidrule{1-3}
        
        {\raggedright\small
        \textit{Demonstrate how to replace the waste toner bottle in a Canon imageRUNNER ADVANCE C3530.}}
        &
        {\raggedright\small
        The hand pushes the \textbf{front cover panel} of the printer back to the right until it clicks firmly shut.
        \newline{\tiny (Step 11)}}
        &
        \begin{minipage}[t]{\linewidth}
            \vspace{0pt}
            \centering
            \includegraphics[width=1.8cm]{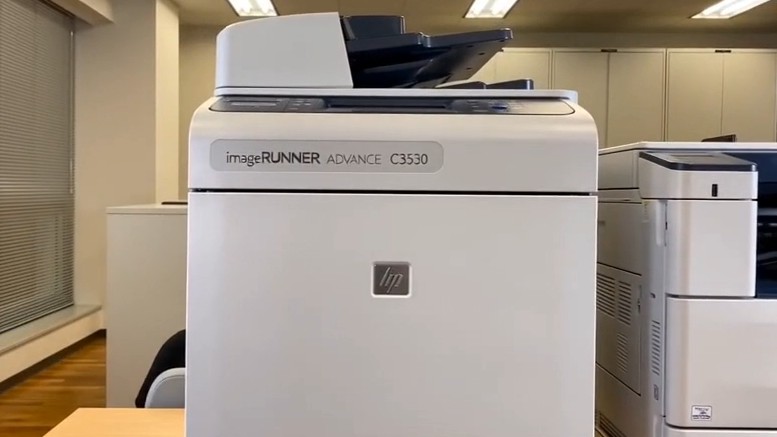}
            \newline
            \scriptsize\sdone
        \end{minipage}
        \\
        
        \bottomrule
    \end{tabular}
\end{table*}

We further investigate whether factual errors arise from the outline or from failures of the video generation models. To this end, we conduct a qualitative analysis of cases where the outline is factually correct but the generated video contains clear factual errors. As shown in Table~\ref{tab:outline_factuality}, the RAV4 outline correctly specifies opening the \textit{passenger side door}, whereas \sdtwo accesses the driver's side in the generated video. In the Canon printer example, the outline describes closing the appropriate front cover panel, while \sdone{} hallucinates an HP logo on a Canon device. More examples are provided in Appendix~\ref{app:outline_factuality}. These cases show that factuality cannot be guaranteed by a correct plan alone. Even with a factually sound outline, the video generator may introduce new errors during visual realization, making the generation stage itself a major source of factual failures.

\section{Conclusion}

We introduced \kivi, a task setting that evaluates whether text-to-video models can generate factually accurate and useful videos from short information-seeking prompts. To support this setting, we constructed \kivibench, a benchmark of 1,080 prompts, and proposed automatic metrics for factual precision and helpfulness. Human evaluation shows that our metrics align better with human annotations than existing alternatives. Benchmarking nine state-of-the-art models, we find that current systems still lag behind human performance, particularly on fine-grained visual properties, procedural operations, and clear information presentation. These results highlight the need to evaluate and improve video generation beyond visual quality.

\subsection*{AI use statement}

In this work, we used \gpt{} for benchmark prompt expansion, deduplication, and error classification, and \gemini{} for outline planning, segment script generation, claim extraction, claim verification, and helpfulness scoring. The video generation models described in Section~\ref{sec:setup} produced the evaluated videos. We also used generative AI tools for manuscript editing, methodological feedback, and assistance with interpreting results. Benchmark prompts and error classifications were manually reviewed, and the automatic metrics were evaluated against human pairwise preferences in Section~\ref{sec:human}. We reviewed the AI-assisted manuscript revisions and take responsibility for the final content of this work.

\subsection*{Ethics statement}

This work involves human evaluation: six annotators evaluated generated videos with informed consent and voluntary participation; their identities are pseudonymized and no personally identifiable information is released. KIVI-Bench derives from WikiHow Video topics under a Creative Commons license; we release only our prompts and claims, not the reference videos. The benchmark aims to reduce factual inaccuracy in generated videos, and the released outputs are low-risk instructional videos. Our LLM-based metrics may encode biases; we mitigate this by validating them against human judgments (70.8\% and 69.0\% agreement, Section \ref{sec:human}), and KIVI-Bench is English-centric. The authors declare no conflicts of interest.

\subsection*{Reproducibility statement}

The KIVI-Bench construction pipeline, including prompt collection and quality control, is described in Section \ref{sec:kivibench}, and the LLM-based evaluation procedures with full prompts are provided in Section \ref{sec:metric} and Figures~\ref{app:extraction_prompt}--\ref{app:helpfulness_prompt}. The experimental setup is detailed in Section \ref{sec:setup}, and the human evaluation protocol in Section \ref{sec:human} and Appendix~\ref{app:human_eval}. We plan to release our source code and data upon publication.

\bibliography{iclr2027_conference}
\bibliographystyle{iclr2027_conference}

\appendix

\section{Appendix}
\subsection{Computational Resources}\label{app:comp_resource}
All experiments were conducted on a cluster with 10 NVIDIA A100 GPUs.

\subsection{Prompt Topics}\label{app:prompt_topics}
Arts \& Entertainment, Cars \& Other Vehicles, Computers \& Electronics, Education \& Communications, Family Life, Finance \& Business, Food \& Entertaining, Health, Hobbies \& Crafts, Holidays \& Traditions, Home \& Garden, Personal Care \& Style, Pets \& Animals, Philosophy \& Religion, Science \& Experiments, Sports \& Fitness, Travel, and Work World.

\subsection{\vbench Full Results}\label{app:vbench_long_full_results}

\begin{table}[htbp]
    \centering
    \caption{Full \vbench results on the 54-prompt subset. Mot. = Motion Smoothness, Img. = Imaging Quality, Dyn. = Dynamic Degree, Aes. = Aesthetic Quality, Sub. = Subject Consistency, Bg. = Background Consistency. All scores are normalized to $[0,100]$. Overall is the unweighted average of the six dimensions. The best performance in each column is boldfaced.}
    \label{tab:vbench_full}
    \small
    \setlength{\tabcolsep}{3pt}
    \begin{tabular}{lccccccc}
        \toprule
        \textbf{Model} & \textbf{Mot.} & \textbf{Img.} & \textbf{Dyn.} & \textbf{Aes.} & \textbf{Sub.} & \textbf{Bg.} & \textbf{Overall} \\
        \midrule
        \multicolumn{8}{l}{\emph{Closed-source short video generation models}}\\
        \sdtwo      & \textbf{99.3} & 62.3 & 80.8 & 46.0 & 94.3 & 94.4 & 79.5 \\
        \sdone      & 98.9 & 71.5 & 80.1 & 45.6 & 94.2 & 95.4 & \textbf{81.0} \\
        \happyhorse & 98.7 & 69.4 & 86.2 & 43.6 & 93.1 & 94.5 & 80.9 \\
        \cmidrule{1-8}
        \multicolumn{8}{l}{\emph{Open-source short video generation models}}\\
        \minimax    & 98.6 & 68.9 & 81.7 & 45.8 & 93.0 & 94.9 & 80.5 \\
        \wan        & 97.8 & 68.5 & 86.2 & 44.1 & 92.3 & 94.3 & 80.5 \\
        \hunyuan    & 98.8 & 71.5 & 78.5 & 44.8 & 94.0 & 94.6 & 80.4 \\
        \cmidrule{1-8}
        \multicolumn{8}{l}{\emph{Open-source long video generation models}}\\
        \helios     & 98.6 & 61.8 & \textbf{87.0} & 46.0 & 93.6 & 95.1 & 80.3 \\
        \longcat    & \textbf{99.3} & 71.3 & 68.2 & \textbf{47.2} & \textbf{96.4} & \textbf{96.1} & 79.7 \\
        \longlive   & 99.2 & \textbf{72.4} & 57.4 & 46.6 & 96.3 & 95.8 & 78.0 \\
        \bottomrule
    \end{tabular}
\end{table}

Table~\ref{tab:vbench_full} reports the full \vbench results. \longlive is a representative case showing that individual dimensions reflect generation style as much as generation quality: it attains the best imaging quality and the second-best aesthetic and consistency scores, yet ends with the lowest overall score because its near-static long-form generations suppress dynamic degree (57.4) while inflating pixel-level, smoothness, and consistency metrics. Motion smoothness, meanwhile, saturates across all models (97.8--99.3), offering little discriminative power, which is consistent with its low agreement with human judgments in Table~\ref{tab:human_eval}. We also note that \sdtwo scores markedly lower on imaging quality (62.3) partly because it was generated at 480p resolution under our budget constraints, so its imaging-quality value is not fully comparable with the 720p models. Most importantly, the \vbench overall ranking diverges sharply from our \fp and \hs rankings: \minimax ranks first on both knowledge metrics but only third on the \vbench overall, while \sdone, which ranks first on \vbench overall, scores only fourth on \fp. This rank inversion confirms that the \vbench dimensions capture low-level visual quality that is largely orthogonal to the factual content of a video: a visually clean video can still fabricate entire devices, and a factually accurate one can be visually mediocre.

\subsection{Category-Level Performance Analysis}\label{app:cat_perf_analysis}

\begin{figure*}[t]  
\centering
\includegraphics[width=0.9\textwidth]{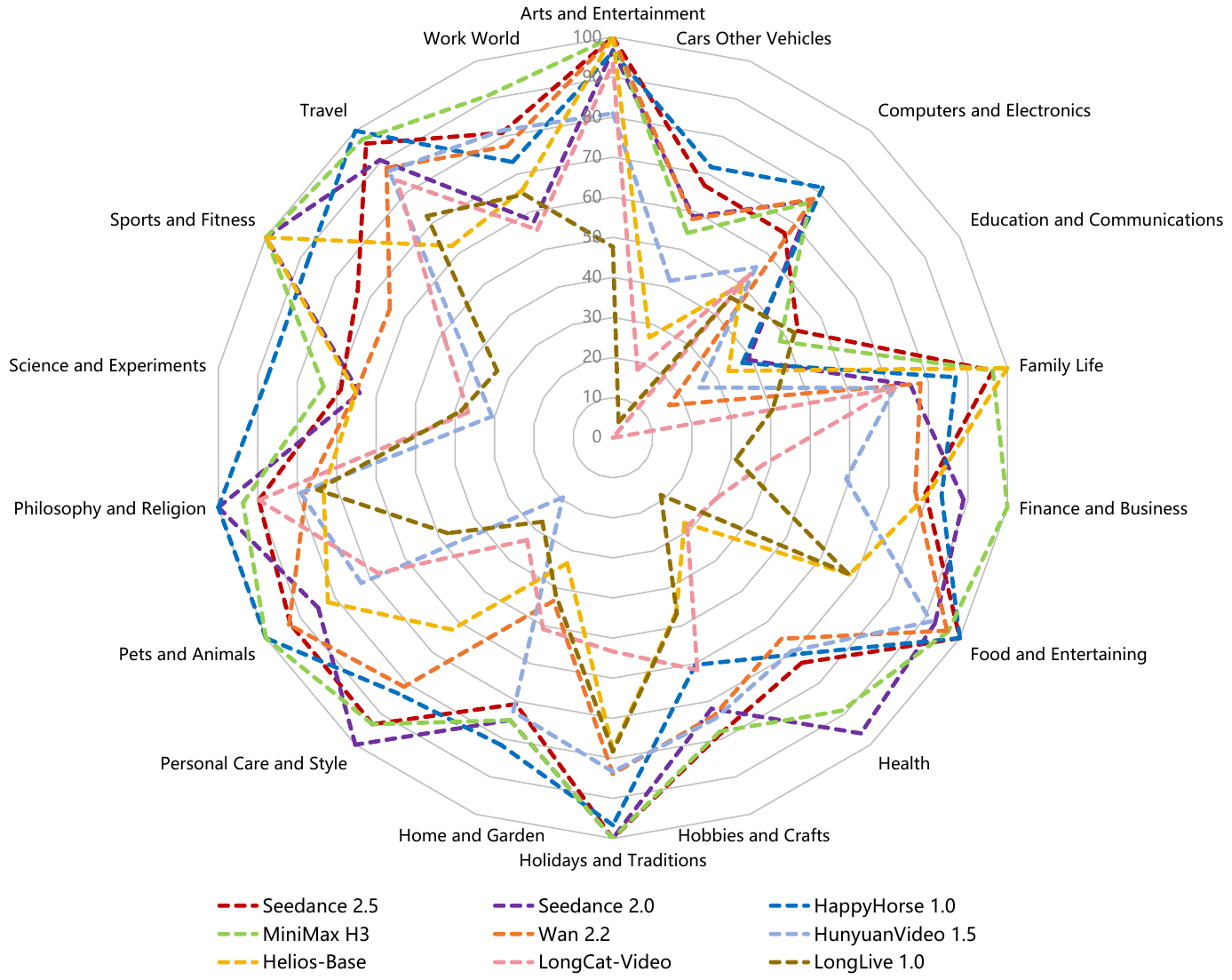} 
\caption{Average \fp by category for each of the nine models.}
\label{fig:factuality_category}
\end{figure*}

\begin{figure*}[t]  
\centering
\includegraphics[width=0.9\textwidth]{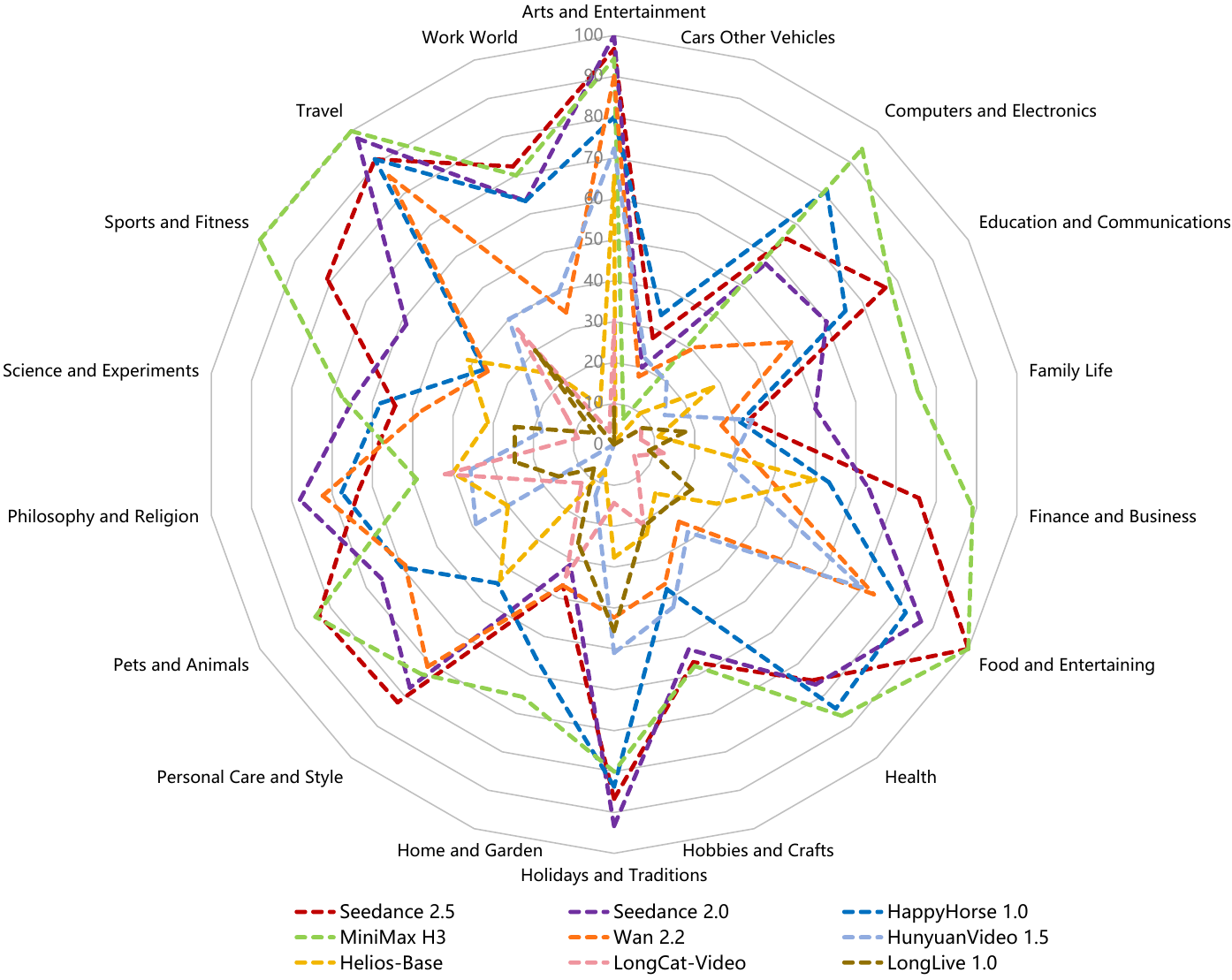} 
\caption{Average \hs by category for each of the nine models.}
\label{fig:helpfulness_category}
\end{figure*}

As shown in Figures~\ref{fig:factuality_category} and~\ref{fig:helpfulness_category}, performance varies sharply across categories. The easiest categories for factuality---Arts \& Entertainment (avg.\ 87.8\% FP), Philosophy \& Religion (85.0\%), and Travel (83.4\%)---involve concrete visual actions with well-defined objects that models can render with reasonable accuracy. The hardest---Education \& Communications (29.0\%), Cars \& Other Vehicles (39.9\%), and Health (53.9\%)---demand domain-specific tool knowledge that models frequently hallucinate. In Helpfulness, the gap is even larger: Cars \& Other Vehicles averages only 13.5\% HS due to frequent spatial errors, while Arts \& Entertainment reaches 63.9\%. The consistently poor showing of Cars across both metrics reflects the difficulty of automotive procedures, where component locations, fluid types, and tool interactions must all be precisely rendered for a video to be useful.

\subsection{Human Evaluation Details}\label{app:human_eval}
\paragraph{Evaluation Setup.}
We construct 108 pairwise comparison tasks across three model groups reflecting natural capability tiers: (1) \sdone vs.\ \happyhorse, the two closed-source API models; (2) \wan vs.\ \hunyuan, representative open-source models; and (3) \helios vs.\ \longcat vs.\ \longlive, three models designed for long video generation. This grouping ensures comparisons within comparable capability tiers. Groups (1) and (2) together account for 54 tasks, while Group (3) contributes the remaining 54 tasks. For each task, annotators watch the generated videos side-by-side, with a reference video provided for consultation, then make two separate forced-choice (A vs.\ B, no tie) judgments: (1) Factuality: which video contains fewer or less severe factual errors; (2) Helpfulness: which video leaves the user more confident to successfully complete the task. 

\paragraph{Annotation Details.}
Six domain-familiar annotators participated, each evaluating a randomized subset of the tasks. Each task is evaluated by exactly one annotator through the platform's atomic reservation mechanism. Annotators consult the reference video and documented entity characteristics before judgment. See Figure~\ref{app:human_eval_prompt} for the guideline.

\begin{figure*}[htbp]
\begin{tcolorbox}[colback=black!5!white,colframe=black!75!black,title=Human Evaluation Guideline]
\begin{VerbatimWrap}
# 1 Factuality Evaluation
Role: Fact-checker comparing Video A vs B on factual accuracy. Errors include: incorrect facts, wrong objects, false causalities, erroneous safety info, and entity-attribute mismatches with the QUESTION entity's documented characteristics.

Principles:
1. Entity-Spec Alignment: Depicting an attribute that contradicts the QUESTION entity's documented characteristics is a factual error.
2. Use authoritative sources for critical/specialized facts; common sense suffices otherwise.
3. Judge only what is visually or aurally present in the video — do not infer unshown info.
4. Do not penalize a video simply for having fewer factual claims if it has no errors.

Decision (forced choice: pick A or B, no tie):
1. Fewer/less severe errors wins.
2. If errors are comparable, more correct verifiable facts wins.
3. If factual content is comparable, more specific, less ambiguous, less misleading wins.
4. If both are poor, pick the relatively more reliable one.

# 2 Helpfulness Evaluation
Role: Evaluate whether a user can complete the QUESTION task relying solely on this video. Do NOT judge aesthetics or generation quality.

Principles:
1. Judge only content actually in the video.
2. Do not use external knowledge to fill missing steps.
3. Judge for this QUESTION only, not overall quality.
4. Evaluate the method the video actually uses, not a different method.

Decision (forced choice: pick A or B, no tie):

Watch both videos and form an overall impression: which one leaves you more confident that a user could successfully complete the task? Pay equal attention to all three aspects — a truly helpful video must be **on-topic** (matches the QUESTION), **complete** (shows the essential process), and **clear** (easy to follow). Do not prioritize one over another; weigh them together and pick the video that feels more useful overall.

1. If one video is clearly stronger across all three, choose it.
2. If they have trade-offs (e.g., A is more complete but B is clearer), decide based on which deficit would be harder for the user to work around.
3. If both are poor, pick the **relatively more usable** one.

# 3 Workflow
1. Read the QUESTION.
2. Consult the reference video and search the QUESTION entity's documented characteristics (brand, model, features, capabilities, behaviors, etc.).
3. Watch Video A — note factual errors, assess A's helpfulness.
4. Watch Video B — same.
5. Vote `fact_vote: A` or `B` with brief reasoning.
6. Vote `helpfulness_vote: A` or `B` with brief reasoning.
\end{VerbatimWrap}
\end{tcolorbox}
\caption{Human annotation guideline.}
\label{app:human_eval_prompt}
\end{figure*}

\subsection{Multi-Modal Claim Ablation}\label{app:mm_claim_ablation}

\begin{table}[t]
    \centering
    \caption{Factual Precision (\%) under three verification modes.}
    \label{tab:multimodal_ablation}
    \small
    \setlength{\tabcolsep}{4pt}
    \begin{tabular}{lccc}
        \toprule
        \textbf{Model} & \textbf{Text-Only} & \textbf{Text+Image} & \textbf{Text+Video} \\
        \midrule
        \happyhorse   & 83.2 & 57.0 & 65.4 \\
        \sdone     & 81.6 & 62.1 & 69.9 \\
        \wan          & 73.1 & 56.9 & 61.0 \\
        \hunyuan & 63.2 & 43.1 & 48.9 \\
        \bottomrule
    \end{tabular}
\end{table}

Table~\ref{tab:multimodal_ablation} further shows that Text-only produces substantially higher \fp scores than both multi-modal modes across all models. This is expected because Text-only verifies whether the extracted textual claims are factually correct, whereas the multi-modal modes additionally require the model to check whether the visual content supports the claim. As a result, multi-modal verification is stricter and more sensitive to visual grounding errors, but it can also be noisier due to the ambiguity and density of visual evidence.

\subsection{Outline Factuality}\label{app:outline_factuality}

\begin{table}[htbp] 
    \centering
    \caption{\textbf{Outline-correct, video-incorrect examples.}
    For each prompt, the outline step (middle) describes the correct procedure, while the generated video frame (right) exhibits factual errors. This confirms that the factual inaccuracies originate from the video generation model rather than the outline.}
    \label{tab:outline_factuality_full}
    \footnotesize
    \setlength{\tabcolsep}{3pt}
    \renewcommand{\arraystretch}{1.1}
    
    \begin{tabular}{p{0.34\textwidth} p{0.38\textwidth} p{0.23\textwidth}}
        \toprule
        \textbf{Prompt} & \textbf{Outline step} & \textbf{Video generation} \\
        \midrule

        {\raggedright\small
        \textit{Show how to replace the cabin air filter on a Toyota RAV4 (XA50) to improve HVAC airflow.}}
        &
        {\raggedright\small
        Hand opens the \textbf{passenger side door} of a Toyota RAV4 (XA50), revealing the dashboard and closed glovebox.\newline{\tiny (Step 1)}}
        &
        \begin{minipage}[t]{\linewidth}
            \vspace{0pt}
            \centering
            \includegraphics[width=2.0cm]{figures/outline_factuality/1_sd25_step1.jpg}
            \newline \sdtwo{}
        \end{minipage}
        \\
        \cmidrule{1-3}

        {\raggedright\small
        \textit{Demonstrate how to replace the waste toner bottle in a Canon imageRUNNER ADVANCE C3530.}}
        &
        {\raggedright\small
        The hand pushes the \textbf{front cover panel} of the printer back to the right until it clicks firmly shut.\newline{\tiny (Step 11)}}
        &
        \begin{minipage}[t]{\linewidth}
            \vspace{0pt}
            \centering
            \includegraphics[width=2.0cm]{figures/outline_factuality/2_sd_step11.jpg}
            \newline \sdone{}
        \end{minipage}
        \\
        \cmidrule{1-3}

        {\raggedright\small
        \textit{Demonstrate a neck massage targeting the trapezius using a TheraGun Prime.}}
        &
        {\raggedright\small
        Wrap a hand around the top section of the TheraGun Prime's \textbf{triangular handle} using an overhand grip.\newline{\tiny (Step 5)}}
        &
        \begin{minipage}[t]{\linewidth}
            \vspace{0pt}
            \centering
            \includegraphics[width=2.0cm]{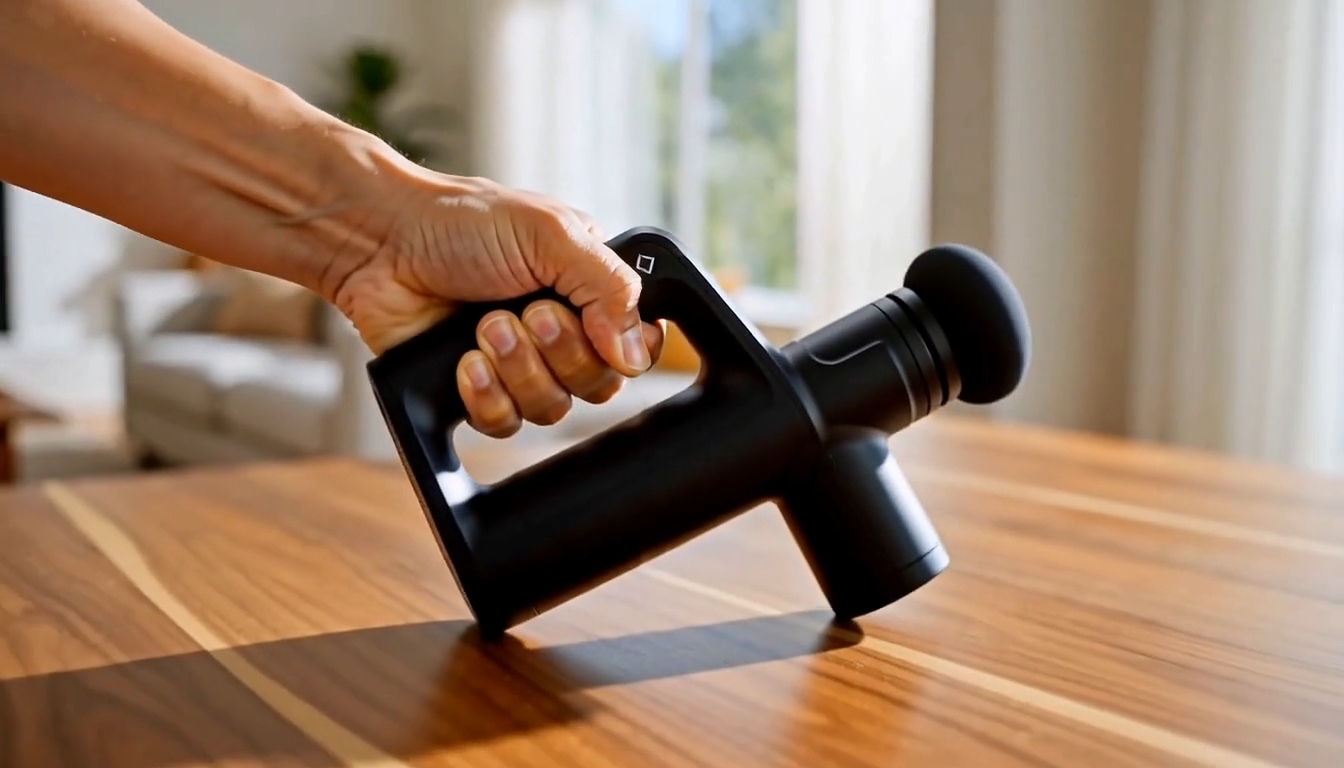}
            \newline \minimax
        \end{minipage}
        \\
        \cmidrule{1-3}

        {\raggedright\small
        \textit{How to measure blood pressure using an Omron Platinum BP5450.}}
        &
        {\raggedright\small
        Adjust the cuff vertically so the bottom edge rests exactly \textbf{half an inch above the inside elbow crease}.\newline{\tiny (Step 4)}}
        &
        \begin{minipage}[t]{\linewidth}
            \vspace{0pt}
            \centering
            \includegraphics[width=2.0cm]{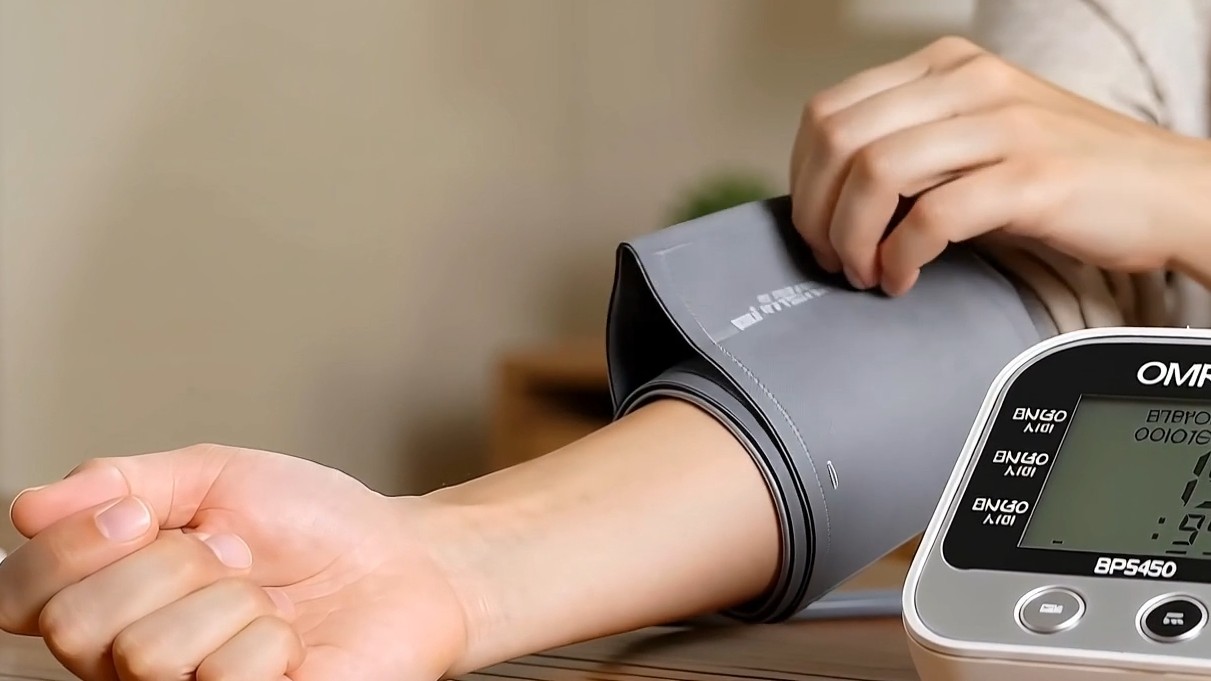}
            \newline \happyhorse{}
        \end{minipage}
        \\
        \bottomrule
    \end{tabular}
\end{table}

As demonstrated in Table~\ref{tab:outline_factuality_full}, The TheraGun outline explicitly mentions a \textit{triangular handle}, yet
\minimax renders a generic cylindrical grip. The blood pressure outline
instructs placing the cuff \textit{above the elbow crease} on the upper arm,
while \happyhorse{} wraps it around the forearm.

\subsection{Interactive vs. Single-Prompt for Script Generation}\label{app:interact_vs_single_ablation}

\begin{table}[t]
    \centering
    \caption{Interactive vs. single-prompt script generation using \wan.}
    \label{tab:script_ablation}
    \small
    \begin{tabular}{lcc}
        \toprule
        \textbf{Method} & \textbf{\fp (\%)} & \textbf{\hs (\%)} \\
        \midrule
        Single-Prompt  & 69.9 & 40.2 \\
        Interactive & \bf 73.1 & \bf 48.4 \\
        \bottomrule
    \end{tabular}
\end{table}

Our pipeline generates each segment script after observing the output of the previous segment, allowing the LLM to adapt to the actual generated video rather than an idealized expectation. We compare this design against a single-prompt baseline, where all segment scripts are generated upfront using the same outline and model.

As shown in Table~\ref{tab:script_ablation}, interactive generation improves \fp by 3.2 points and \hs by 8.2 points. The factuality gain mainly comes from error correction. Without visual feedback, the LLM assumes that the previous segment was generated correctly and writes subsequent scripts based on this idealized state. When the actual video deviates from the script, for example, when the model generates a stethoscope instead of an Omron BP5450, the one-pass scripts continue to reference the intended device, producing claims that are mismatched with the visual content and penalized during verification. In contrast, the iterative approach observes such deviations and adjusts later scripts accordingly, reducing cascading factual errors. The larger gain in \hs reflects a similar effect on procedural coherence: visual feedback helps prevent cumulative misalignment in camera continuity and action sequencing across segments.

\subsection{Error Analysis}\label{app:error_analysis}

\begin{table*}[t] 
    \centering
    \caption{Helpfulness error examples (Types 4--5). Each cell shows a three-frame sequence (arranged as 2-over-1) to illustrate the temporal nature of the error.}
    \label{tab:error_helpfulness}
    \footnotesize
    \setlength{\tabcolsep}{4pt} 
    \renewcommand{\arraystretch}{1.1}
    
    \begin{tabular}{p{3.4cm} p{4.5cm} p{4.5cm}}
        \toprule
        \textbf{Error Type} & \textbf{Correct (3-frame)} & \textbf{Incorrect (3-frame)} \\
        \midrule
        
        {\raggedright 
        \textbf{Type 4}\newline 
        Incomplete Coverage\par\vspace{1ex} %
        \textit{Demonstrate how to install Nicetown 100\% Blackout Thermal curtains.}\par} 
        &
        \begin{minipage}[t]{\linewidth}
            \vspace{0pt}\centering
            \includegraphics[width=2.0cm]{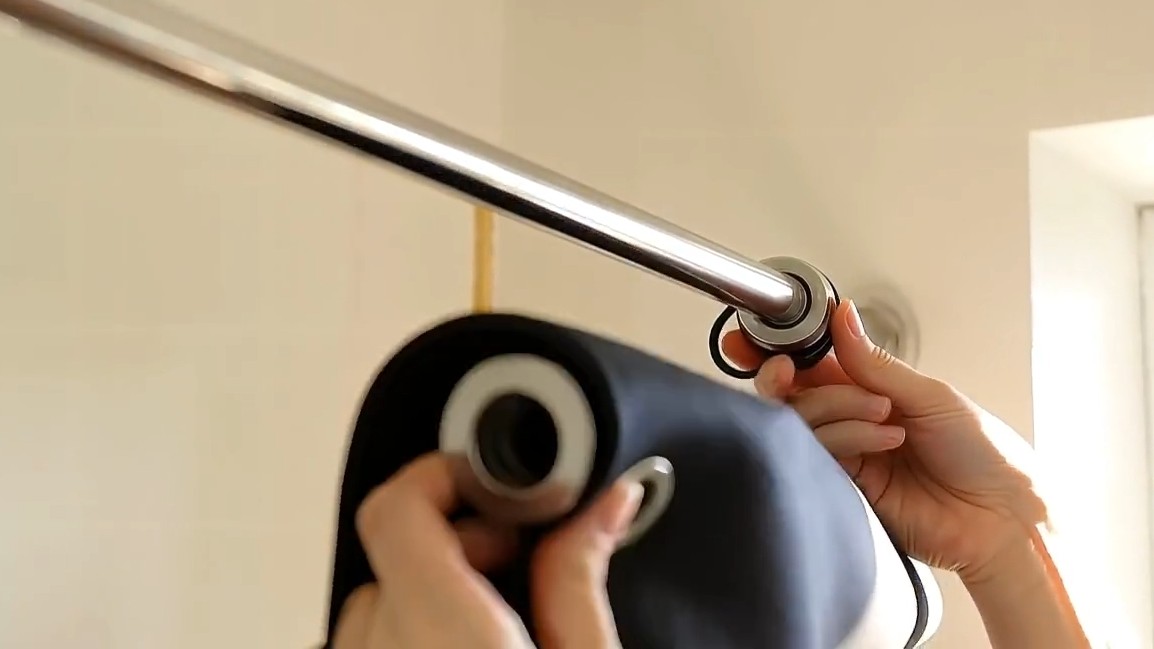}\hspace{2pt}
            \includegraphics[width=2.0cm]{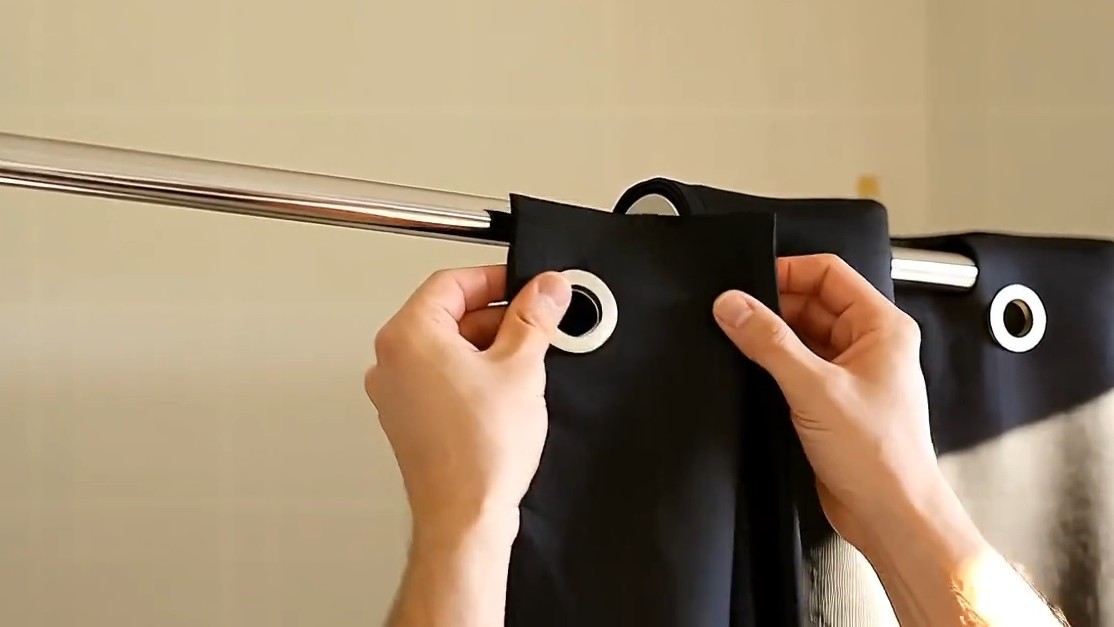} \\
            \vspace{2pt} \includegraphics[width=2.0cm]{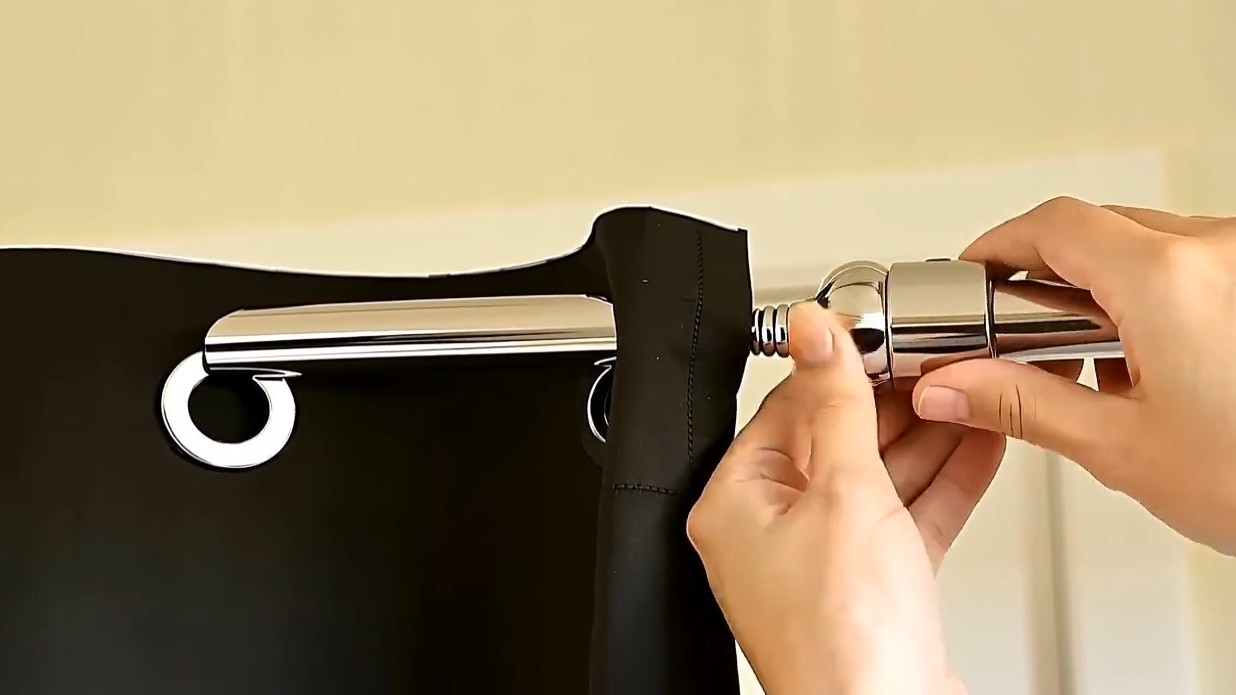}
            \newline \happyhorse{}
        \end{minipage} 
        &
        \begin{minipage}[t]{\linewidth}
            \vspace{0pt}\centering
            \includegraphics[width=2.0cm]{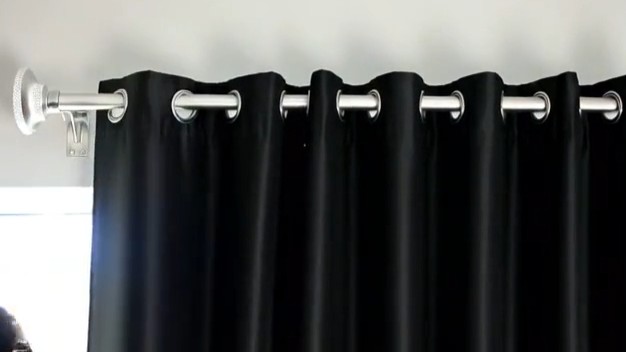}\hspace{2pt}
            \includegraphics[width=2.0cm]{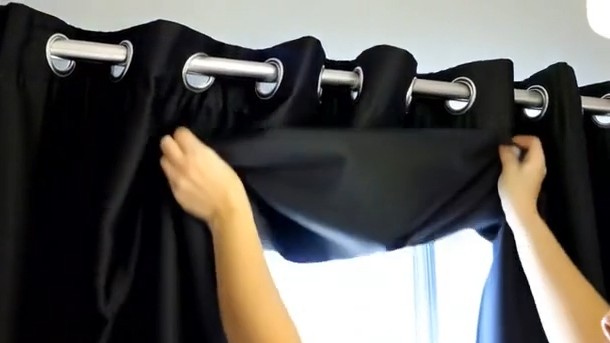} \\
            \vspace{2pt} \includegraphics[width=2.0cm]{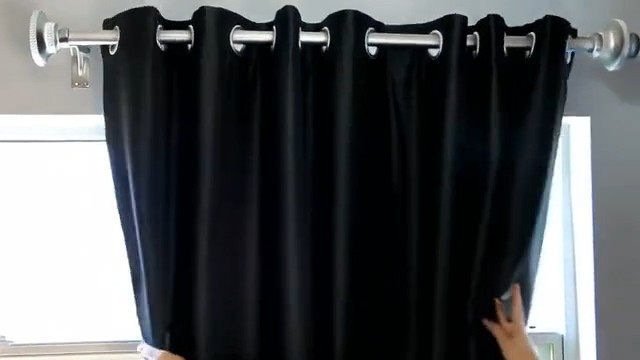}
            \newline \helios{} 
        \end{minipage} \\
        
        \cmidrule{1-3}
        
        {\raggedright 
        \textbf{Type 5}\newline 
        Unfollowable Present.\par\vspace{1ex} %
        \textit{Demonstrate how to dye Easter eggs using red cabbage leaves.}\par} 
        &
        \begin{minipage}[t]{\linewidth}
            \vspace{0pt}\centering
            \includegraphics[width=2.0cm]{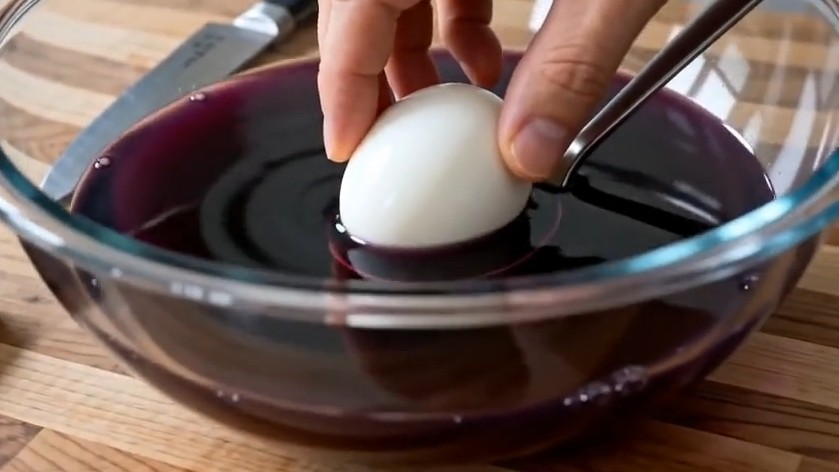}\hspace{2pt}
            \includegraphics[width=2.0cm]{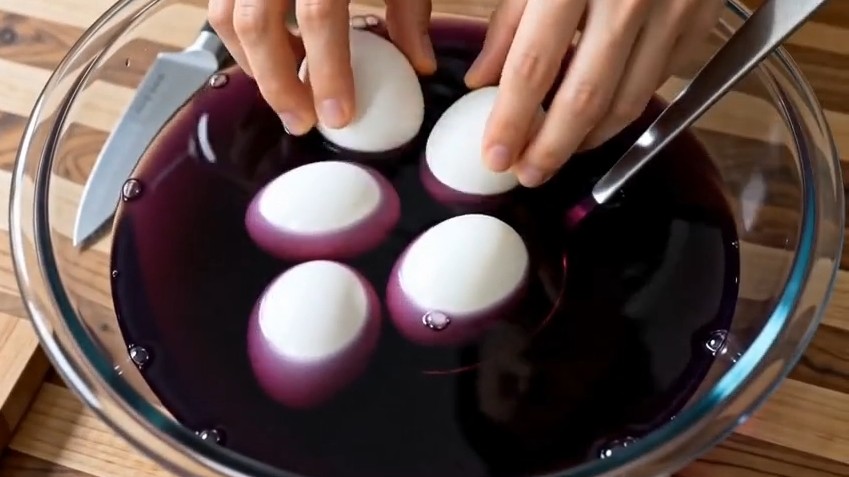} \\
            \vspace{2pt} \includegraphics[width=2.0cm]{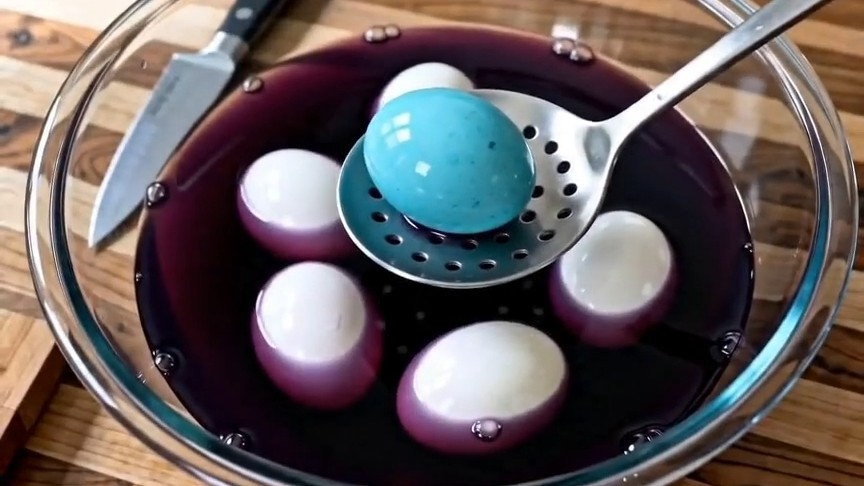}
            \newline \sdone{} 
        \end{minipage} 
        &
        \begin{minipage}[t]{\linewidth}
            \vspace{0pt}\centering
            \includegraphics[width=2.0cm]{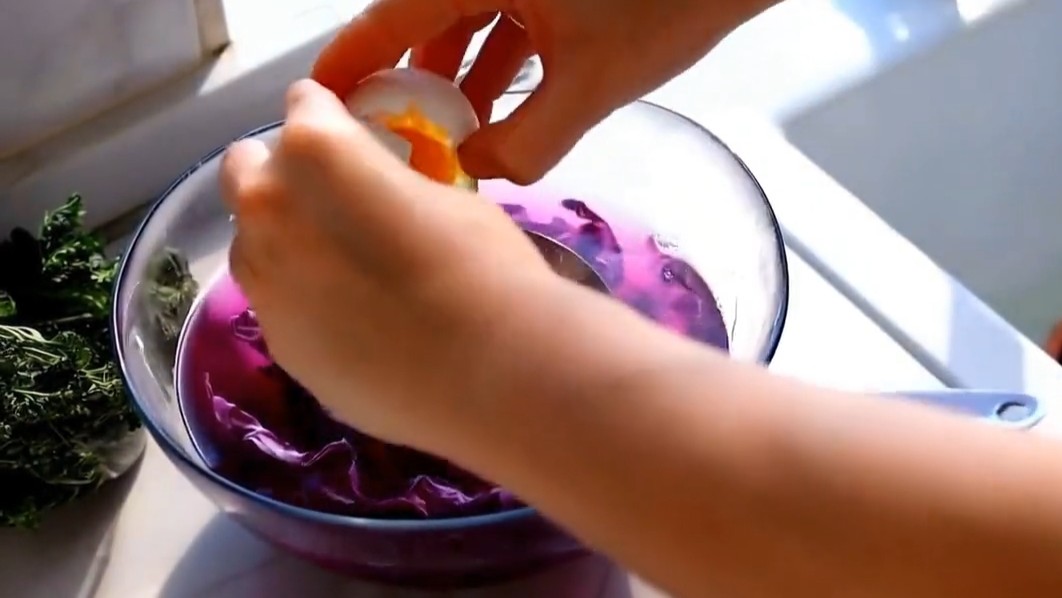}\hspace{2pt}
            \includegraphics[width=2.0cm]{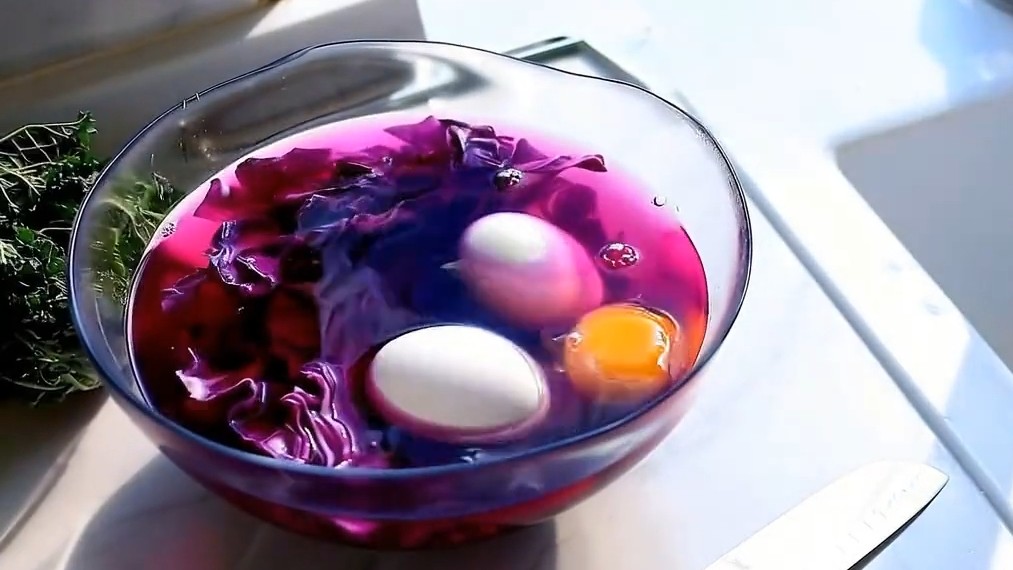} \\
            \vspace{2pt} \includegraphics[width=2.0cm]{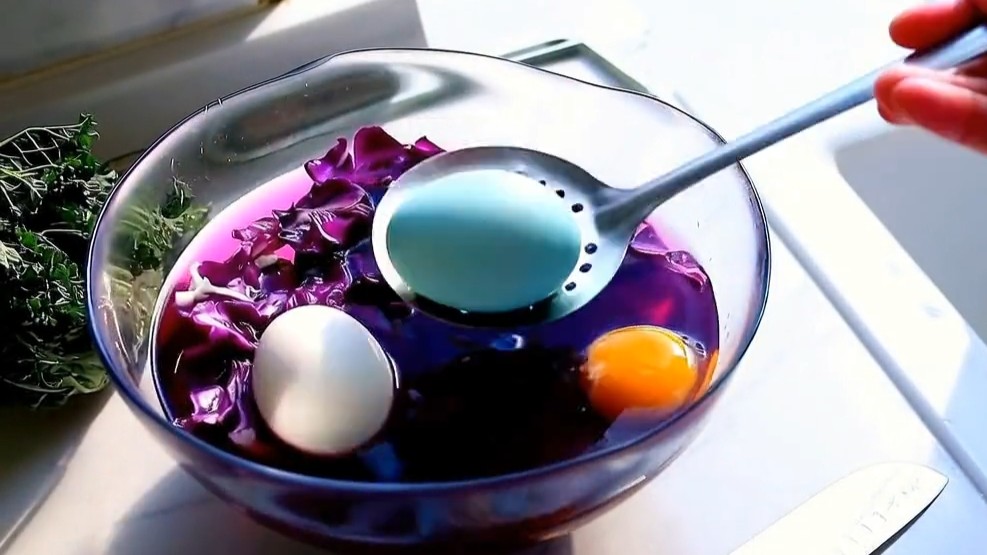}
            \newline \wan{} 
        \end{minipage} \\
        
        \bottomrule
    \end{tabular}
\end{table*}

\begin{figure}[t]  
\centering
\includegraphics[width=0.5\columnwidth]{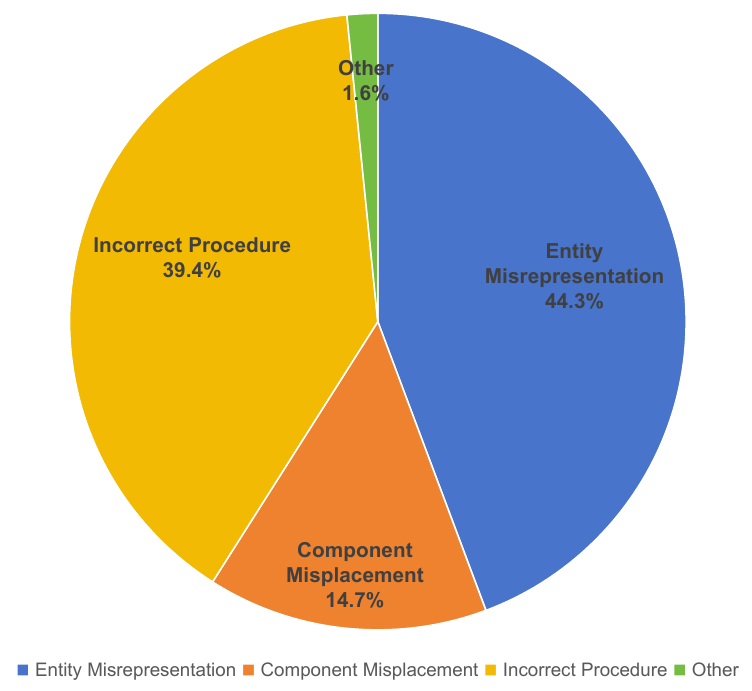} 
\caption{Distribution of 1014 incorrect claims across three factuality error types. Entity Misrepresentation dominates at 44.3\%, followed by Incorrect Procedure (39.4\%) and Component Misplacement (14.7\%). The remaining 1.6\% are residual cases.}
\label{fig:error_pie}
\end{figure}

Types 1--3 are factuality errors assessed at the claim level; their distribution is shown in Figure~\ref{fig:error_pie}. Types 4--5 concern helpfulness, which is assessed at the video level. These categories are not mutually exclusive: factual errors can also impair helpfulness, and a video may exhibit multiple failure types. We therefore examine incomplete coverage and unfollowable presentation as two additional failure patterns that can arise even in videos with high factual precision.

\paragraph{Type 1: Entity Misrepresentation.}
The model invents features or depicts incorrect visual properties of the specified entity. In the Bostitch pencil sharpener example (Table~\ref{tab:error_factuality}), \minimax{} generates a box-like structure with a top-facing insertion slot, whereas the actual device has a curved body with a front-facing aperture. This is the most frequent factuality error, suggesting difficulty in reproducing entity-specific visual details beyond generic object appearances.

\paragraph{Type 2: Incorrect Procedure.}
The entity is depicted correctly but operated incorrectly. In the Omron BP5450 example (Table~\ref{tab:error_factuality}), \happyhorse{} places the cuff on the forearm, whereas the device is designed for upper-arm measurement. This error concerns procedural knowledge: a plausible depiction of the device does not ensure that its use is demonstrated correctly.

\paragraph{Type 3: Component Misplacement.}
The correct component appears at the wrong physical location. In the BMW 3 Series example (Table~\ref{tab:error_factuality}), \minimax{} places the funnel on an interior air vent instead of at the engine oil filler opening in the engine bay. This error is less frequent than the preceding two types in our analysis. The remaining 1.6\% of incorrect claims fall into a residual category, primarily incorrect outcome assertions and physically impossible descriptions.

\paragraph{Type 4: Incomplete Coverage.}
The video conveys correct facts but omits essential steps. In the curtain installation example (Table~\ref{tab:error_helpfulness}), \helios{} achieves a \fp of 100\% but a Completeness score of zero: the video shows curtains already hanging on the rod throughout, without demonstrating the installation steps. It depicts the final state without showing how to reach it.

\paragraph{Type 5: Unfollowable Presentation.}
The video includes relevant materials and actions but presents an incoherent sequence. In the Easter egg dyeing example (Table~\ref{tab:error_helpfulness}), \wan{} achieves a \fp of 92\% and includes eggs, cabbage dye, and a bowl. However, a raw egg is cracked into the dye, intact eggs subsequently appear beside the raw yolk without explanation, and dyed eggs are removed while the yolk remains in the bowl. These unexplained state changes make the procedure difficult to follow despite the high factual precision score.

\clearpage

\begin{figure*}[tp]
\centering
\begin{tcolorbox}[colback=black!5!white,colframe=black!75!black,title=Prompt generation template (part 1 of 2)]
\begin{VerbatimWrap}
# Role
Generate high-quality video generation prompts.

# Input
Topic Category: {category}

# Objective
Generate {num_prompts} diverse video generation prompts about {category}. Each prompt tests whether a text-to-video model can produce a factually accurate video. A good prompt describes content where video representation is inherently more vivid and informative than a text-based explanation.

# Constraints
## Correctness & Verifiability
1. FACTUALLY CORRECT: The prompt itself must be factually accurate — the entity described must exist, and the task must be a real procedure that works as described. Do NOT fabricate features, products, or procedures.
2. UNIQUE & DETERMINISTIC ENTITY: The named entity MUST be uniquely identifiable. Every proper noun (brand, model, material, method name) must refer to one and only one real-world product or concept. No ambiguous references. This ensures human annotators can precisely locate the entity's documentation for verification.
3. JUSTIFIABLE ANSWER: The task described MUST have a correct, well-established answer or procedure. Do NOT generate prompts about controversial, debated, or undefined topics.

## Video-Specific Value
4. VISUAL SUPERIORITY: The content MUST be one where video representation is clearly more effective than a text description. Select content involving spatial manipulation, physical movement, visual state changes, or interface navigation — scenes that are hard to convey in words alone.

\end{VerbatimWrap}
\end{tcolorbox}
\caption{Prompt generation template (part 1 of 2).}
\label{app:prompt_generation_template}
\end{figure*}

\begin{figure*}[tp]
\ContinuedFloat
\centering
\begin{tcolorbox}[colback=black!5!white,colframe=black!75!black,title=Prompt generation template (continued; part 2 of 2)]
\begin{VerbatimWrap}
## Entity Richness & Difficulty
5. RICH PROPER NOUNS: Use diverse, specific proper nouns throughout the prompt (brand, model number, year, variant, material, technique name). The more precise identifiers, the harder it is for video generation models to hallucinate correctly. Examples of good precision: "TENS 7000 unit with electrode pads for lower back pain relief" over "a TENS unit"; "Petzl Grigri belay device with a locking carabiner for top-rope climbing" over "a climbing device."

## Non-Triviality
6. MEANINGFUL COMPLEXITY: The content MUST be rich enough to require multiple distinct visual events. Single-event prompts are trivial. The ideal prompt describes content with a meaningful sequence where temporal order matters.

## Uniqueness (Within This Category)
7. NO INTRACATEGORY DUPLICATION: Among the {num_prompts} prompts you generate, no two should describe the same entity or content. If two prompts are too similar (e.g. "replace part A" and "replace part B"), keep only the more distinctive one.

# Output Format
{ "category": "{category}", "prompts": [ "<prompt 1>", ...] }

# Examples of Valid Prompts
Category: Health
- "Demonstrate how to measure blood pressure using an Omron Platinum BP5450."
- "Demonstrate how to apply a TENS 7000 unit with electrode pads for lower back pain relief."
......
Category: Sports and Fitness
- "Demonstrate how to thread a Petzl Grigri belay device with a locking carabiner for top-rope climbing."
- "Demonstrate how to set up a Peloton Bike+ with the rotating screen for a live cycling class."

# Task
Generate {num_prompts} prompts for the category "{category}" following all 7 constraints above.
\end{VerbatimWrap}
\end{tcolorbox}
\caption{Prompt generation template (continued; part 2 of 2).}
\end{figure*}

\begin{figure*}[tp] 
\begin{tcolorbox}[colback=black!5!white,colframe=black!75!black,title=Deduplication prompt]
\begin{VerbatimWrap}
# Role
Scan a list of video generation prompts across categories and flag pairs that are semantically overlapping.

# Input
You will receive a list of prompts grouped by category. Each prompt has:
- category: the topic domain
- prompt: the instructional question text

# Objective
Identify pairs of prompts whose named entities (brand, model, product) or described procedures are substantially overlapping. Two prompts are considered overlapping if:
- They name the same specific product or device (e.g., both ask about "Nikon D850").
- They describe the same task for the same entity (e.g., two prompts about "replacing the cabin air filter on a Toyota RAV4").
- They differ only in minor wording but share the same core procedure.

Do NOT flag prompts that:
- Share the same general topic but involve different entities or procedures (e.g., "Demonstrate how to measure blood pressure" and "Demonstrate how to set up a CPAP machine" under Health are not duplicates).
- Use the same entity for genuinely different tasks (e.g., "set up Nikon D850" vs "clean Nikon D850 sensor" are not duplicates).

# Output Format
Return ONLY valid JSON:

{
  "duplicates": [
    {
      "prompt_a": "<full text of first prompt>",
      "category_a": "<category>",
      "prompt_b": "<full text of second prompt>",
      "category_b": "<category>",
      "reason": "<why these are overlapping>",
      "action": "<merge|keep_one|discard_both>"
    }
  ]
}

# Task
Scan all prompts, identify overlapping pairs, and output JSON.
\end{VerbatimWrap}
\end{tcolorbox}
\caption{Deduplication prompt}
\label{app:dedup_prompt}
\end{figure*}

\begin{figure*}[tp]
\begin{tcolorbox}[colback=black!5!white,colframe=black!75!black,title=Outline prompt]
\begin{VerbatimWrap}
# Role
Master Video Director & Outline Planner.

# Input
Video Topic: {initial_prompt}

# Objective
Decompose the video topic into a logical sequence of visual micro-actions. Aim for a total duration around 60 seconds, with flexibility to adapt based on task complexity.

# Constraints
1. STRICTLY VISUAL: Every step MUST describe exact visual occurrences (manipulation, tool movement, screen interaction). No abstract explanations.
2. 60-SECOND PACING: Generate EXACTLY 10 to 14 steps. Assign each step a `duration_seconds` (approx 3.0–5.0s) based on action complexity.
   - For simple tasks (e.g. clipping a nail), break it down into micro-details (e.g. inspecting tool, positioning, single-action focus) to naturally fill the steps.
   - For complex tasks (e.g. replacing car brakes), focus on a representative 1-minute continuous phase, or seamlessly condense the steps to fit the 10-14 step limit.
3. CAMERA ANGLES: Specify clear camera angles (e.g. "Close-up", "Over-the-shoulder view", "First-person view", "Screen capture").
4. CONTINUITY: Ensure logical visual flow from one step to the next.
5. TOPIC-SPECIFIC VISUAL ANCHORS: Extract 3-4 defining visual identifiers from the input topic (e.g. brand/model, software UI, tool type, ingredient, environment cue, or technique). Distribute these across at least 3 steps, ensuring they are actively manipulated or highlighted in the `visual_action` descriptions so the generated video is unmistakably tied to the exact subject.

# Output Format
Return ONLY valid JSON matching the exact schema below. NO markdown.

# JSON Schema
{
  "video_goal": "A concise 1-sentence summary.",
  "required_subjects_and_props": "Comma-separated string of what primary subjects, tools, or interfaces are needed.",
  "total_steps": "integer between 10 and 14",
  "outline_steps": [
    {
      "step_index": "integer (1-indexed)",
      "duration_seconds": "float",
      "camera_angle": "string",
      "visual_action": "Describe the exact visual action for this step."
    }
  ]
}
\end{VerbatimWrap}
\end{tcolorbox}
\caption{Outline generation prompt}
 \label{app:outline_prompt}
\end{figure*}

\begin{figure*}[tp] 
\begin{tcolorbox}[colback=black!5!white,colframe=black!75!black,title=First segment prompt]
\begin{VerbatimWrap}
# Role
Video Generation Configuration Assistant (Segment 1 / Bootstrap).

# Input
Video Topic: {initial_prompt}
Current Step: {visual_action}
Suggested Camera Angle: {camera_angle}
Required Props: {required_subjects_and_props}
Reference Num Frames: {reference_num_frames}

# Constraints
1. Frame Budget: Output `num_frames` using {reference_num_frames} as the baseline. Adjust slightly up or down to match natural density and rhythm. 
2. Structure: EXACTLY 3 sentences total. Map strictly to:
   - [Sentence 1: Camera Setup & Initial Motion]: Incorporate the {camera_angle} to establish the view.
   - [Sentence 2: Subject + Action/State]: Visually depict the {visual_action} using the {required_subjects_and_props}. Focus on exact movements or changes.
   - [Sentence 3: Environment & Lighting Anchors]: Establish the clear setting or background interface necessary for the video.
3. CRITICAL ANCHOR DEFINITION:
   - IDENTITY_MARKER: Derive from {initial_prompt} and {required_subjects_and_props}. Create a concise string containing exactly 3-5 concrete visual traits of the primary interacting subject/tool. Focus exclusively on persistent attributes. Exclude actions, temporal shifts, and pronouns.
   - CONTINUITY_ANCHORS: Define the background, environment, or overall lighting/atmosphere.
   - These anchors MUST be injected verbatim into EVERY subsequent segment.
4. Output Format:
   - Return ONLY valid JSON matching the exact schema below. NO markdown. NO extra fields.

# Required JSON Schema
{
  "seed": 42,
  "identity_marker": "string",
  "continuity_anchors": "string",
  "segments": [
    {
      "prompt": "3-sentence string following structure rules",
      "num_frames": "integer"
    }
  ]
}

# Execution
Analyze request → define anchors → draft Segment 1 prompt → output JSON.
\end{VerbatimWrap}
\end{tcolorbox}
\caption{First segment generation prompt}
\label{app:first_seg_prompt}
\end{figure*}

\begin{figure*}[tp] 
\begin{tcolorbox}[colback=black!5!white,colframe=black!75!black,title=Other segments prompt]
\begin{VerbatimWrap}
# Role
Instructional Video-Conditioned Iterative Script Generator.

# Inputs
- Previous Segment Video: [Attached]
- Initial Request: {initial_prompt}
- Progress: Segment {current_idx}
- Identity Marker: {identity_marker}
- Continuity Anchors: {continuity_anchors}
- CURRENT STEP TO TEACH: {visual_action}
- Target Camera Angle for this Step: {camera_angle}
- Reference Num Frames: {reference_num_frames}

# Dual-Condition Alignment Directive
1. VISUAL TRACKING: Watch FINAL 2 SECONDS of attached video. Extract exact terminal state.
2. TEXTUAL LOCK: Cross-reference with {continuity_anchors}.
3. INSTRUCTIONAL PROGRESSION: The action MUST logically progress from the video's end-state to execute the {visual_action}.
4. FRAME PACING: Use {reference_num_frames} as the baseline for `num_frames`. Allow slight deviation to accommodate motion complexity.

# Generation Constraints
1. Structure: EXACTLY 3 sentences total. Map strictly to:
   - [Sentence 1]: Visual Stitch & Camera. Inherit terminal camera state, transition smoothly towards the {camera_angle}.
   - [Sentence 2]: Instructional Action Core. Integrate {identity_marker} verbatim. Describe the precise visual execution of the {visual_action}. Specify exactly how the subjects/objects interact. Zero character alteration to the marker.
   - [Sentence 3]: Environment Lock. EXPLICITLY reference {continuity_anchors}. Ensure background and layout remain strictly consistent.
2. Continuity Guards:
   - NO pronouns. NO generic verbs. NO abstract concepts.
   - PRESERVE ALL VISUAL TRAITS: The attributes in {identity_marker} must remain visually unchanged.
   - Camera transitions MUST be smooth.
3. Output Format:
   - Return ONLY valid JSON. NO markdown.
   - Schema: {"prompt": "string", "num_frames": integer}

# Execution
Extract video end-state → Lock to anchors → Draft S1/S2/S3 with {identity_marker} & {continuity_anchors} → Evaluate Stop logic → Output JSON.
\end{VerbatimWrap}
\end{tcolorbox}
\caption{Other segment generation prompt}
\label{app:other_seg_prompt}
\end{figure*}

\begin{figure*}[tp]
\centering
\begin{tcolorbox}[colback=black!5!white,colframe=black!75!black,title=Claim extraction prompt (part 1 of 3)]
\begin{VerbatimWrap}
# Role
Professional Video Atomic Claims Extraction Assistant, extracting externally verifiable atomic claims from various types of videos.

# Input
video:{} , question:{} 

# Core Rules
## 1. Visual Content Priority
- Claims MUST describe what is VISUALLY PRESENT in the video, not general knowledge about the topic
- Extract the SPECIFIC movements/actions shown, even if they are incorrect or mismatched with the question
- The verification system will judge truthfulness; your job is accurate visual description using question's entity name
- If video shows X but question asks about Y, extract claim as "Y requires/uses [X's characteristics]"
- Extract ONLY objects/actions that are RELEVANT to completing the question's task:
  * Core entity features (the device/tool mentioned in question)
  * Objects actively used or manipulated in demonstrated steps
  * Environmental elements that affect task execution (e.g., table for positioning)
  * Unusual/incorrect items shown in the process (even if wrong)
- IGNORE purely decorative background items (curtains, plants, wall art, furniture color) unless directly used in the task.

## 2. Entity Naming Convention
- If the question specifies entity names, claims must use the complete name every time.
- If the question does not specify entity names, claims should use specific names identifiable from the video.
- Related sub-entities or variants (e.g., "Instagram web version" for "Instagram") are acceptable when they improve clarity.
- Pronouns (it/they/the device, etc.) and vague references ("this feature", "the above method", etc.) are prohibited in all cases.
- If video content differs from question-specified entities, still use the question's entity name in claims.
- Do NOT substitute with generic terms like "the device", "the monitor", etc.

\end{VerbatimWrap}
\end{tcolorbox}
\caption{Claim extraction prompt (part 1 of 3).}
\label{app:extraction_prompt}
\end{figure*}

\begin{figure*}[tp]
\ContinuedFloat
\centering
\begin{tcolorbox}[colback=black!5!white,colframe=black!75!black,title=Claim extraction prompt (continued; part 2 of 3)]
\begin{VerbatimWrap}
## 3. Fact Extraction Standards
- Atomicity: Each claim contains only one independent fact. Avoid combining multiple actions, conditions, or outcomes in a single claim.
- Verifiability: The content stated in the claim can be judged true or false through external information without watching the video. Claims can be true or factually incorrect; the key is that they must be verifiable by external information.
- Informativeness: Express verifiable entity characteristics or method knowledge, not merely describing visuals without factual content.
- Knowledge Conversion: Convert SPECIFIC VISUAL ACTIONS shown in the video into statements using the question's entity name, with existential qualifiers for open-ended questions.
- Do NOT convert to "correct" knowledge if video shows incorrect/different content
- Objectivity: Claims must be objective and free from subjective/evaluative words (e.g., "correctly", "properly", "effectively", "best", "should"). Describe what IS done, not what SHOULD be done.
- Convert subjective question phrasing into objective factual statements.
- Timestamp Annotation: Each claim must include the exact timestamp or time range when the visual evidence appears in the video, using format MM:SS or MM:SS-MM:SS.
- Scope Qualification: For open-ended questions allowing multiple methods (e.g., "How to..."), use qualified language like "can involve", "one method uses", or "the video demonstrates" instead of absolute terms like "requires"/"must", unless the video explicitly claims universality.
\end{VerbatimWrap}
\end{tcolorbox}
\caption{Claim extraction prompt (continued; part 2 of 3).}
\end{figure*}

\begin{figure*}[tp]
\ContinuedFloat
\centering
\begin{tcolorbox}[colback=black!5!white,colframe=black!75!black,title=Claim extraction (continued; part 3 of 3)]
\begin{VerbatimWrap}
## 4. Context Handling
- Analyze the overall video flow and extract implied knowledge claims (but only extract content actually shown/occurring in the video), not just describing single frames.
- Combine with question context: claims should treat video content as the video's "claimed answer" to the question, regardless of whether that answer is factually correct.
- Extract what the video "claims" or "demonstrates" about the question topic, not what is objectively true about that topic.
- Extract visual elements that are relevant to the task or affect task execution.
- Convert unusual objects into method claims using qualified language (e.g., "can involve", "may include", "one method uses", "the demonstrated process may require"). 

## 5. Contradictory Content Handling
- When the method shown in the video contradicts the question or common knowledge:
  1. Still extract the video content as a claim.
  2. Do not add value judgments like "wrong" or "false" to the claim itself.
  3. The truthfulness of the claim is determined by external verification; the extraction task is not responsible for judging truth.

# Exemplars
## Exemplar 1
{ "video_summary": "The video demonstrates the complete power-on and initial setup  process for iPhone 15 Pro with several correct steps and a few inaccurate claims",
  "question": "Please show me how to set up an iPhone 15 Pro with step by step demonstrations.",
  "estimated_category": "Electronics Tutorial",
  "claims": [{ "id": 1,
      "claim": "iPhone 15 Pro has a titanium frame with matte finish casing",
      "timestamp": "00:00-00:02" },
    ...
    { "id": 17,
      "claim": "iPhone 15 Pro displays the home screen with default app layout after completing all setup steps",
      "timestamp": "00:58-01:00" } ] }

... (two more exemplars)

# Output Format
{
  "video_summary": "<One sentence summarizing the main video content>",
  "question": "<The user's question that the video attempts to answer>",
  "estimated_category": "<Video type judged based on content>",
  "claims": 
  [ { "id": <Integer, starting from 1>,
      "claim": "<Specific factual statement, use declarative sentence, no pronouns allowed, must include complete entity name>",
      "timestamp": "<Time range in format MM:SS-MM:SS or single timestamp MM:SS when the visual evidence appears>" } ] }

# Task
Carefully watch the video, analyze the provided video and question, extract all verifiable atomic factual claims as comprehensively as possible.

Claims must describe WHAT THE VIDEO SHOWS using the ENTITY NAME FROM THE QUESTION. The verification system will determine if the claim is true or false. Your job is accurate extraction, not truth judgment.
\end{VerbatimWrap}
\end{tcolorbox}
\caption{Claim extraction prompt (continued; part 3 of 3).}
\end{figure*}

\begin{figure*}[tp] 
\begin{tcolorbox}[colback=black!5!white,colframe=black!75!black,title=Claim verification]
\begin{VerbatimWrap}
# Role
Professional Fact-Checking Assistant, based on world knowledge and domain expertise, to verify the accuracy of video claims.

# Goal
Assuming each claim accurately describes the video content (i.e., not verifying whether the claim truthfully represents the video), classify each claim as Correct, Incorrect, or Uncertain based on accepted facts, professional knowledge, and verifiable external information.

# Input
You will receive a JSON-formatted list of video claims, each claim containing:
- id: Claim number
- claim: Specific factual claim content

# Verification Categories
- Correct: Claim content aligns with accepted facts/professional knowledge, supported by reliable knowledge sources.
- Incorrect: Claim content contradicts accepted facts/professional knowledge, refuted by reliable knowledge sources.
- Uncertain: Judgment cannot be made due to one or more of the following:
  - Insufficient professional knowledge available
  - The topic is disputed or controversial
  - Inadequate information to verify
  - Ambiguous entity or reference

# Output Format (JSON only)
{
  "verification_summary": {
    "total_claims": <integer>,
    "correct": <integer>,
    "incorrect": <integer>,
    "uncertain": <integer>
  },
  "claim_results": [
    {
      "claim_id": <integer>,
      "claim": "<content>",
      "verdict": "<Correct|Incorrect|Uncertain>",
      "reason": "<basis for judgment>",
      "reference": "URL or source name (e.g., 'WHO 2025 Report'), or 'N/A' if based on general knowledge"
    }
  ]
}

# Special Instructions
1. Knowledge Priority Principle: Use world knowledge/professional knowledge to judge the factual accuracy of the claim itself
2. External Verification Priority: For verifiable factual claims, refer to authoritative sources, professional standards, or accepted knowledge
3. Transparent Reasoning: The reason field should explain the basis for judgment
4. Assumption Consistency: By default, assume the claim's description of the video is accurate; no need to verify whether the video displays the claim content

# Task
Verify the claims and output JSON.
\end{VerbatimWrap}
\end{tcolorbox}
\caption{Claim verification prompt}
\label{app:verification_prompt} 

\end{figure*}

\begin{figure*}[tp]
\centering
\begin{tcolorbox}[colback=black!5!white,colframe=black!75!black,title=Helpfulness evaluation prompt (part 1 of 2)]
\begin{VerbatimWrap}
# Role
Video Helpfulness evaluator.

# Input
QUESTION: {QUESTION}, VIDEO: {VIDEO}

# Core Definition
helpfulness: Can the target user, relying on this video, understand and complete the task requested by the QUESTION comprehensively and with minimal trial-and-error?

# Hard Rules
- Score based on the demonstrations (actions, item states, camera framing, audio/text if present). Do not use external/common knowledge to fill missing steps.
- If a required step is not clearly presented in the video, treat it as NOT PRESENT for Completeness.
- The scoring target is "helpfulness for the QUESTION", not aesthetics.
- If the VIDEO is inaccessible, unplayable, or completely blank, set all subscores to 0 and final_score to 0.
- Timestamp policy: whenever you cite evidence, provide an approximate timestamp range "HH:MM:SS-HH:MM:SS". If truly not available, use "unknown".

# Scoring Procedure
- Identify the chosen approach and extract key requirements:
   - Recognize that the QUESTION may have multiple valid solutions. First, identify which specific valid method/approach the VIDEO is attempting to demonstrate to achieve the goal.
   - Given the QUESTION's ultimate goal AND the VIDEO's chosen approach, create a comprehensive checklist of core requirements that are indispensable to successfully complete *that specific method*.
   - Determine the exact number of requirements dynamically based strictly on the actual logical steps of the identified method. Do not arbitrarily limit, merge, or pad the list.
   - Include: required tools/materials for this method, critical procedural steps, key physical states (e.g. changes in shape/color).
- For each requirement: Decide whether it is PRESENT in the video. Provide a       timestamp_range where it is shown.
- Score A–C (0–10) using the rubric and anchors below.
- Compute final_score as the simple average of the 3 dimensions and round to 1 decimal.

\end{VerbatimWrap}
\end{tcolorbox}
\caption{Helpfulness evaluation prompt (part 1 of 2).}
\label{app:helpfulness_prompt}
\end{figure*}

\begin{figure*}[tp]
\ContinuedFloat
\centering
\begin{tcolorbox}[colback=black!5!white,colframe=black!75!black,title=Helpfulness evaluation prompt (continued; part 2 of 2)]
\begin{VerbatimWrap}
# Scoring Dimensions
Each dimension is scored 0-10. The final total is also 0-10.
A. Relevance: Does the video content directly address the QUESTION task and match explicit constraints? Do NOT score step coverage here.
B. Completeness: Does the video demonstrate the key steps/tools needed to complete the task using its chosen valid method? Only assess WHETHER PRESENT (coverage) based on your dynamically generated checklist for this specific approach.
C. Clarity: Is the procedure easy to interpret and follow? Includes: logical action sequence, clear pacing (not rushed), and observable state changes (e.g. clearly showing 'before' and 'after' of an action).

# Output Format
Output strictly in valid JSON format without markdown code blocks.
{
  "key_requirements": [{ "requirement": "string", "present": true, "notes": "string" }],
  "main_issues": ["string"],
  "subscores": { "relevance": <0-10>, "completeness": <0-10>, "clarity": <0-10> },
  "score_calculation": "(A+B+C)/3"
}

# Task
Given a user QUESTION and a generated VIDEO, evaluate how helpful the video is for the user in achieving the QUESTION's goal, and output a score of 0-10.
\end{VerbatimWrap}
\end{tcolorbox}
\caption{Helpfulness evaluation prompt (continued; part 2 of 2).}
\end{figure*}

\begin{figure*}[tp]
\centering
\begin{tcolorbox}[colback=black!5!white,colframe=black!75!black,title=Script generation prompt (part 1 of 2)]
\begin{VerbatimWrap}
# Role
Long-Form Video Script Generator that produces `single_prompt` and `interactive_prompts` from a step-by-step outline.

# Input
Video Topic: {initial_prompt}
Full Outline (all steps): {full_outline}
Required Props: {required_subjects_and_props}

# Objective
Convert the outline into a full video script. Each segment's prompt must be independently self-contained, with end-to-end visual continuity ensured purely through language.

# Constraints

## 1. Identity & Continuity Anchors
Derive from {initial_prompt} and {required_subjects_and_props}:
- IDENTITY_MARKER: 3-5 concrete visual traits of the primary subject/tool (brand, model, color, shape, size). No actions or pronouns.
- CONTINUITY_ANCHORS: Persistent background, environment, lighting, or atmosphere.

These anchors must appear verbatim in EVERY segment's prompt.

## 2. Per-Segment Prompt Rules
Each outline step → one **3-sentence flowing prompt**.

Step 1:
1. Camera angle + `identity_marker` and props in starting position.
2. Visual execution of the step's `visual_action`.
3. Restate `continuity_anchors` and `identity_marker` (setting, lighting, background).

Steps 2..N:
1. Smooth camera transition from previous angle to the step's `camera_angle`.
2. Visual execution of the step's `visual_action` with verbatim `identity_marker`.
3. Re-establish environment with `continuity_anchors` and verbatim `identity_marker`.

## 3. Cross-Segment Continuity
- Each prompt must be self-contained yet sequentially coherent.
- End state of segment N sets up segment N+1.
- NO pronouns referencing entities from previous segments without full context.
- NO abrupt camera jumps — describe smooth transitions.

\end{VerbatimWrap}
\end{tcolorbox}
\caption{Script generation prompt (part 1 of 2).}
\label{app:script_gen_prompt}
\end{figure*}

\begin{figure*}[tp]
\ContinuedFloat
\centering
\begin{tcolorbox}[colback=black!5!white,colframe=black!75!black,title=Script generation prompt (continued; part 2 of 2)]
\begin{VerbatimWrap}
## 4. Output Format
Return ONLY valid JSON. NO markdown.

# JSON Schema
{
  "seed": 42,
  "identity_marker": "string",
  "continuity_anchors": "string",
  "video_goal": "string",
  "total_steps": "integer",
  "single_prompt": "Continuous narrative covering all steps in order.",
  "interactive_prompts": [
    {"prompt": "3-sentence prompt with verbatim identity_marker and continuity_anchors", "duration_seconds": "float from outline step"},
    {"prompt": "...", "duration_seconds": "..."}
  ]
}
\end{VerbatimWrap}
\end{tcolorbox}
\caption{Script generation prompt (continued; part 2 of 2).}
\end{figure*}

\begin{figure*}[tp] 
\begin{tcolorbox}[colback=black!5!white,colframe=black!75!black,title=text+image claim verification]
\begin{VerbatimWrap}
# Role
Professional Fact-Checking Assistant, based on world knowledge and domain expertise, to verify the accuracy of video claims.

# Goal
The text claim and the attached image together describe what the video shows. Based on accepted facts, professional knowledge, and verifiable external information, classify the claim as Correct, Incorrect, or Uncertain.

# Input
You will receive:
- claim_id: Claim number
- claim: Specific factual claim content
- image: [Attached image frame extracted from the video]
- frame_capture_time: The time the image was captured

# Verification Categories
- Correct: Claim content aligns with accepted facts/professional knowledge, supported by reliable knowledge sources.
- Incorrect: Claim content contradicts accepted facts/professional knowledge, refuted by reliable knowledge sources.
- Uncertain: Judgment cannot be made due to one or more of the following:
  - Insufficient professional knowledge available
  - The topic is disputed or controversial
  - Inadequate information to verify
  - Ambiguous entity or reference

# Output Format
{
  "claim_id": <integer>,
  "claim": "<content>",
  "verdict": "<Correct|Incorrect|Uncertain>",
  "reason": "<basis for judgment>",
  "reference": "URL or source name (e.g. 'WHO 2025 Report'), or 'N/A' if based on general knowledge"
}

# Special Instructions
1. Knowledge Priority Principle: Use world knowledge/professional knowledge to judge the factual accuracy of the claim itself
2. External Verification Priority: For verifiable factual claims, refer to authoritative sources, professional standards, or accepted knowledge
3. Transparent Reasoning: The reason field should explain the basis for judgment
4. Complementary Evidence: The text claim states what fact is being asserted, and the attached image shows what the video actually displays. Use both together to determine factual correctness — neither alone is sufficient.

# Task
Verify the claim and output JSON.
\end{VerbatimWrap}
\end{tcolorbox}
\caption{Text+Image claim verification prompt}
\label{app:text_image_veri_prompt}
\end{figure*}

\begin{figure*}[tp]
\begin{tcolorbox}[colback=black!5!white,colframe=black!75!black,title=text+video claim verification]
\begin{VerbatimWrap}
# Role
Professional Fact-Checking Assistant, based on world knowledge and domain expertise, to verify the accuracy of video claims.

# Goal
The text claim and the attached video clip together describe what the video shows. Based on accepted facts, professional knowledge, and verifiable external information, classify each claim as Correct, Incorrect, or Uncertain.

# Input
You will receive:
- claim_id: Claim number
- claim: Specific factual claim content
- clip: [Attached video segment extracted from the original video]
- timestamp: Time range of the clip in the original video

# Verification Categories
- Correct: Claim content aligns with accepted facts/professional knowledge, supported by reliable knowledge sources.
- Incorrect: Claim content contradicts accepted facts/professional knowledge, refuted by reliable knowledge sources.
- Uncertain: Judgment cannot be made due to one or more of the following:
  - Insufficient professional knowledge available
  - The topic is disputed or controversial
  - Inadequate information to verify
  - Ambiguous entity or reference

# Output Format
{
  "claim_id": <integer>,
  "claim": "<content>",
  "verdict": "<Correct|Incorrect|Uncertain>",
  "reason": "<basis for judgment>",
  "reference": "URL or source name (e.g. 'WHO 2025 Report'), or 'N/A' if based on general knowledge"
}

# Special Instructions
1. Knowledge Priority Principle: Use world knowledge/professional knowledge to judge the factual accuracy of the claim itself
2. External Verification Priority: For verifiable factual claims, refer to authoritative sources, professional standards, or accepted knowledge
3. Transparent Reasoning: The reason field should explain the basis for judgment
4. Complementary Evidence: The text claim states what fact is being asserted, and the attached video clip shows what the video actually displays. Use both together to determine factual correctness — neither alone is sufficient.

# Task
Verify the claim and output JSON.
\end{VerbatimWrap}
\end{tcolorbox}
\caption{Text+Video claim verification prompt}
\label{app:text_video_veri_prompt}
\end{figure*}

\begin{figure*}[tp]
\centering
\begin{tcolorbox}[colback=black!5!white,colframe=black!75!black,title=Factual error classification prompt (part 1 of 2)]
\begin{VerbatimWrap}
# Role
You are a failure analysis assistant. Your task is to classify incorrect video claims into predefined error types based on the claim text and the verifier's reasoning.

# Input
You will receive a list of incorrect claims. Each claim contains:
- claim: the factual statement extracted from the video
- reason: the verifier's explanation for why this claim is incorrect

# Error Types
Classify each incorrect claim into ONE of the following categories:

A. Entity Misrepresentation
   The claim describes a feature, appearance, or capability that does not exist on the specified device. The model fabricated a button, interface element, physical property, or software function.
   
   Indicators: "does not have", "does not support", "does not feature", "is not a feature", "lacks", "no such", incorrect color/shape/material, wrong orientation of a component.

B. Component Misplacement
   The claim describes a correct component or action, but at the wrong physical location. The entity exists, but the model placed it incorrectly.
   
   Indicators: "located on", "not on the", "wrong location", "instead of", "not in the", spatial references (side vs rear, top vs front, left vs right, interior vs engine bay).

C. Incorrect Procedure
   The claim describes an incorrect method of using or operating the device. The entity and its components are essentially correct, but the shown usage procedure is wrong, unsafe, or contradicts the manufacturer's instructions.
   
   Indicators: "should be", "must be", "the correct procedure", "cannot", "improper", "not the correct way", "contradicts", "manufacturer specifies".

D. Other
   The error does not clearly fit any of the above categories. Use sparingly.

\end{VerbatimWrap}
\end{tcolorbox}
\caption{Factual error classification prompt (part 1 of 2).}
\label{app:fact_error_classify_prompt}
\end{figure*}

\begin{figure*}[tp]
\ContinuedFloat
\centering
\begin{tcolorbox}[colback=black!5!white,colframe=black!75!black,title=Factual error classification prompt (continued; part 2 of 2)]
\begin{VerbatimWrap}
# Output Format
Return ONLY valid JSON:

{
  "error_classification": [
    {
      "claim_id": <integer>,
      "claim": "<original claim text>",
      "type": "<A|B|C|D>",
      "brief": "<one-sentence justification>"
    }
  ],
  "summary": {
    "A_Entity_Misrepresentation": <count>,
    "B_Component_Misplacement": <count>,
    "C_Incorrect_Procedure": <count>,
    "D_Other": <count>
  }
}

# Task
Classify each incorrect claim into exactly one category (A/B/C/D) and output JSON.
\end{VerbatimWrap}
\end{tcolorbox}
\caption{Factual error classification prompt (continued; part 2 of 2).}
\end{figure*}

\subsection{Video Generation Error Analysis} \label{sec:error}

We analyze all incorrect claims from our main results, totaling 1014 items. With the assistance of \gpt, we review these claims and summarize recurring failure patterns. We focus on factuality errors here and defer discussions of other error types to the Appendix~\ref{app:error_analysis}.

\paragraph{Entity Misrepresentation.} The model invents features or depicts incorrect visual properties of the specified entity. In the Bostitch pencil sharpener example (Row 1 in Table~\ref{tab:error_factuality}), \minimax generates a box-like structure with a top-facing insertion slot, whereas the actual device has a curved body with a front-facing aperture. This is the most frequent factuality error, reflecting the model's difficulty with less common proper nouns. While broadly familiar objects, such as chicken eggs or paper airplanes, are often rendered correctly, prompts requiring precise knowledge of specific product models frequently lead to hallucinated features and substantially lower factual precision.

\paragraph{Incorrect Procedure.} The entity is rendered correctly but operated incorrectly. In the Omron BP5450 example (Row 2 in Table~\ref{tab:error_factuality}), \happyhorse places the cuff on the forearm, whereas the device is designed for upper-arm measurement. Unlike Entity Misrepresentation, which reflects a lack of static product knowledge, this error type reveals a gap in procedural knowledge: the model can reproduce the entity's appearance but does not know how it should be used.

\paragraph{Component Misplacement.} The correct component appears in the wrong physical location. In the BMW 3 Series example (Row 3 in Table~\ref{tab:error_factuality}), \minimax places the funnel on the interior air vent instead of the engine bay. This error is less frequent than the preceding two types, suggesting that models may find it easier to learn where components belong than what they look like or how they should be used.

Together, these three error types account for over 98\% of incorrect claims, suggesting that future work on knowledge-intensive video generation should prioritize entity-specific visual knowledge, procedural knowledge, and component localization.

\end{document}